\documentclass[acmsmall]{acmart}
\AtBeginDocument{%
  \providecommand\BibTeX{{%
    \normalfont B\kern-0.5em{\scshape i\kern-0.25em b}\kern-0.8em\TeX}}}

\setcopyright{acmlicensed}
\copyrightyear{2024}
\acmYear{2024}
\acmDOI{XXXXXXX.XXXXXXX}

\acmJournal{CSUR}

\usepackage{booktabs} 
\usepackage{multirow} 

\usepackage{makecell}
\usepackage{longtable}

\begin{document}

\title{Federated Continual Learning for Edge-AI: A Comprehensive Survey}

\author{Zi Wang}
\email{z.wang5@exeter.ac.uk}
\author{Fei Wu}
\email{fw407@exeter.ac.uk}
\author{Feng Yu}
\email{fy274@exeter.ac.uk}
\author{Yurui Zhou}
\email{yz822@exeter.ac.uk}
\author{Jia Hu}
\email{j.hu@exeter.ac.uk}
\author{Geyong Min}
\email{g.min@exeter.ac.uk}
\affiliation{%
  \institution{Department of Computer Science, Faculty of Environment, Science and Economy, University of Exeter}
  \city{Exeter}
  \state{Devon}
  \country{United Kingdom}
  \postcode{EX4 4RN}
}

\renewcommand{\shortauthors}{Wang et al.}

\begin{abstract}

Edge-AI, the convergence of edge computing and artificial intelligence (AI), has become a promising paradigm that enables the deployment of advanced AI models at the network edge, close to users.
In Edge-AI, federated continual learning (FCL) has emerged as an imperative framework, which fuses knowledge from different clients while preserving data privacy and retaining knowledge from previous tasks as it learns new ones.
By so doing, FCL aims to ensure stable and reliable performance of learning models in dynamic and distributed environments.
In this survey, we thoroughly review the state-of-the-art research and present the first comprehensive survey of FCL for Edge-AI. 
We categorize FCL methods based on three task characteristics: federated class continual learning, federated domain continual learning, and federated task continual learning. 
For each category, an in-depth investigation and review of the representative methods are provided, covering background, challenges, problem formalisation, solutions, and limitations.
Besides, existing real-world applications empowered by FCL are reviewed, indicating the current progress and potential of FCL in diverse application domains.
Furthermore, we discuss and highlight several prospective research directions of FCL such as algorithm-hardware co-design for FCL and FCL with foundation models, which could provide insights into the future development and practical deployment of FCL in the era of Edge-AI. 

\end{abstract}






\maketitle

\section{Introduction}

Deep Learning (DL) has emerged as a leading approach in artificial intelligence (AI), with demonstrable efficacy across various scientific fields, including computer vision, natural language processing, and speech recognition \cite{dong2021survey}.
DL utilises artificial neural networks with multiple hidden layers to model high-level abstractions and learn complex patterns and representations from data \cite{zuo2023survey}.
In recent years, the proliferation of DL applications has catalysed advancements in various sectors, exemplified by their role in assisting medical diagnostics \cite{esteva2021deep}, enhancing autonomous driving systems \cite{kuutti2020survey}, and accelerating genomics research \cite{sapoval2022current}.
However, traditional implementations of DL rely on cloud computing systems with centralised servers and data storage, which can raise privacy concerns when collecting user data, incur high communication costs, and increase latency between servers and clients.
To address these challenges, edge computing has emerged as a promising approach, which is a distributed computing paradigm that brings computation and storage closer to data sources, rather than relying on centralised cloud-based data processing.
This paradigm shift can significantly reduce the latency and cost, making it suitable for data-intensive and latency-sensitive AI applications.
Therefore, the convergence of edge computing and AI gives rise to Edge-AI, which aims to 
enable real-time AI applications powered by edge computing.

Edge-AI employs a popular distributed machine learning approach called federated learning (FL) \cite{mcmahan2017communication}, which allows collaborative DL model training across clients while keeping the data localised.
To achieve this, a coordinating server distributes the global model to participating clients, which then train the model using their local data.
By aggregating processed parameters such as gradients rather than raw data from each client on the coordinating server, FL ensures the overall training performance and effectiveness of the global model while complying with data security regulations \cite{zhang2021survey, rieke2020future} such as the General Data Protection Regulation (GDPR) and the Data Protection Act (DPA), addressing growing concerns about user privacy in AI applications.

FL research has mainly focused on model convergence under non-independent and identically distributed (non-IID) data \cite{zhu2021federated}, model aggregation \cite{qi2023model}, security and privacy \cite{mothukuri2021survey}, resource optimisation and incentive mechanisms \cite{khan2020federated}, etc.
Furthermore, most FL works assume that the training dataset of clients is sampled from a static data distribution \cite{wang2024comprehensive} and available from the beginning of the training \cite{criado2022non}.
Whereas, in real-world scenarios, the progressive data collection, the distribution of data, the class of samples, and the number of tasks can change over time, bringing significant challenges to the model adaptability \cite{ma2022continual}.

Recently, continual learning (CL), also known as incremental learning (IL) or lifelong learning (LL), has become an important approach for learning and accumulating knowledge from a continual stream of data \cite{wang2024comprehensive}.
Thus, integrating the concept of CL into the FL framework, known as \textbf{Federated Continual Learning (FCL)}, leverages the strengths of both FL and CL to establish a robust foundation for Edge-AI in dynamic and distributed environments.
However, continual learning from a series of new tasks can cause the model to experience significant performance degradation on previously learned tasks, a phenomenon known as catastrophic forgetting (CF) \cite{wang2024comprehensive}.
FCL deteriorates this problem as FL allows clients to join and leave the learning process arbitrarily.
Furthermore, the heterogeneity of FL clients leads local models to learn diverse knowledge, exacerbating catastrophic forgetting in the global model during the aggregation of these local models.
Recent studies (e.g., \cite{ma2022continual, dong2022federated, zhang2023target, shenaj2023asynchronous}) have proposed solutions to tackle these challenges, giving rise to an emerging research field that is increasingly attracting attention.

\subsection{Related Surveys}

In recent years, comprehensive surveys for FL and CL have been conducted separately \cite{zhang2022federated, ye2023heterogeneous, van2022three, de2021continual, masana2022class}.
\textbf{For federated learning}, 
Zhang \textit{et al.} \cite{zhang2022federated} surveyed FL in the IoT domain and explored FL-empowered IoT applications such as healthcare, smart city, and autonomous driving.
Ye \textit{et al.} \cite{ye2023heterogeneous} focused on the challenges of heterogeneous FL from five perspectives: statistical heterogeneity, model heterogeneity, communication heterogeneity, device heterogeneity and additional challenges.
\textbf{For continual learning},
Van \textit{et al.} \cite{van2022three} reviewed CL methods and summarised three types of CL as a common framework to cross-compare the performances of various methods.
Lange \textit{et al.} \cite{de2021continual} surveyed works on CL for task classification, categorising them into replay-based, regularisation-based, and parameter isolation methods, based on how task information is stored and used throughout the learning process.
Masana \textit{et al.} \cite{masana2022class} focused on class-incremental learning and categorized the existing CL methods for image classification into regularisation, rehearsal, and bias-correction methods.
They also provided an extensive experimental evaluation of those methods for image classification tasks.

These surveys are focused on separate areas of FL and CL.
None of them has systematically investigated the challenges and solutions proposed in the emerging paradigm of FCL, especially in the Edge-AI environments.
Recently, Yang \textit{et al.} \cite{yang2023federated} conducted a survey of FCL from the perspective of knowledge fusion.
They proposed two frameworks, namely synchronous and asynchronous FCL, for addressing the spatial-temporal catastrophic forgetting challenge in FCL with knowledge fusion.
Different from their work, our survey thoroughly investigates and categorizes the existing FCL methods in Edge-AI based on three task characteristics: federated class continual learning, federated domain continual learning, and federated task continual learning.
In sections \ref{sec:fccl}, \ref{sec:fdcl}, and \ref{sec:ftcl}, these taxonomies will be explained in more detail.

\subsection{Aim and Contributions}

This survey aims to comprehensively investigate the state-of-the-art research on FCL to provide an in-depth and consolidated review.
From the perspectives of different task characteristics in FCL, we thoroughly review the background, challenges, and methods of FCL.
Furthermore, we explore existing FCL-empowered applications for Edge-AI. 
This survey also provides an in-depth discussion about future research directions, motivating researchers to address important open challenges in FCL and offering insights that could inspire future advancement in Edge-AI.
\textbf{To the best of our knowledge, this paper is the first comprehensive survey of federated continual learning for Edge-AI.} 

The main contributions of this survey are summarised as follows:
\begin{itemize}
    \item We present a comprehensive review and clear taxonomy of the state-of-the-art FCL research based on different task characteristics: federated class continual learning, federated domain continual learning, and federated task continual learning, including a large number of papers in this rapidly expanding research field.
    The taxonomy, definitions, challenges, and advantages and disadvantages of the representative methods are thoroughly discussed.
    \item We provide a review and summary of current real-world applications empowered by FCL, such as intelligent transportation systems, intelligent medical systems, IoT, and digital twins, highlighting the versatility and potential of FCL for making real-world impact.
    \item We deliberate upon and posit several open research challenges including the lack of universal benchmarks, explainability, algorithm-hardware co-design, and FCL with foundation models, while proposing prospective directions that could inspire the research community to advance the field of FCL for its rapid development and wide deployment in the era of Edge-AI.
\end{itemize}

\begin{figure}[!ht]
  \centering
  \includegraphics[width=1\linewidth]{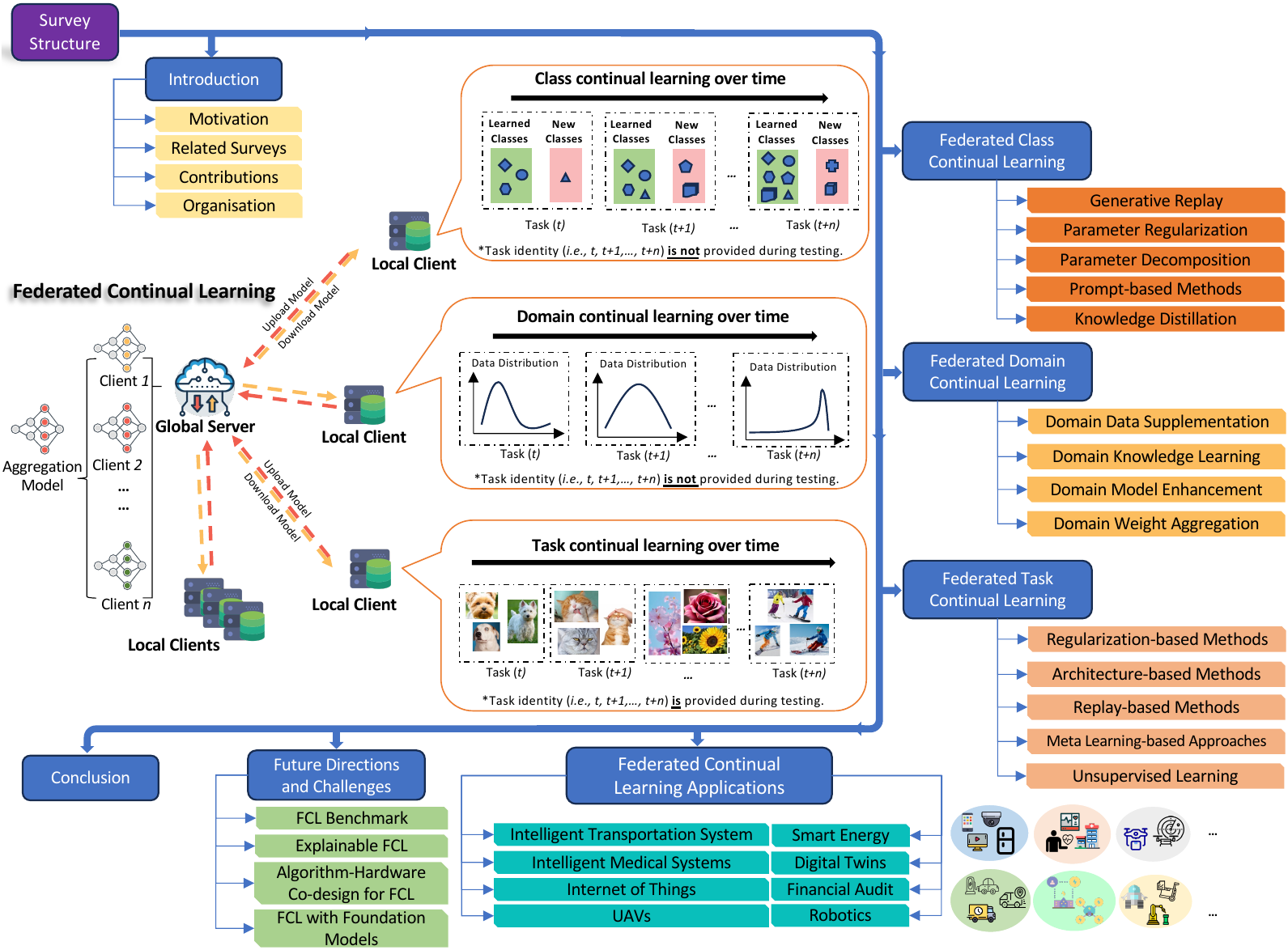}
  \caption{An overview of our federated continual learning survey}
  \label{fig:survey overview}
\end{figure}

\subsection{Survey Organisation}

The overview of the survey is shown in Fig. \ref{fig:survey overview}, and the remainder of this paper is structured as follows.
Section \ref{sec:fccl} first provides a detailed definition of dynamically adding new classes in FCL. 
Then, four types of approaches for this problem are categorized, and we elucidate the interrelationships among these approaches.
Section \ref{sec:fdcl} analyses four types of solutions to the problem of domain drift in FCL.
Section \ref{sec:ftcl} analyses current popular approaches in federated task continual learning.
Section \ref{sec:apps} investigates various applications empowered by FCL.
Section \ref{sec:future} discusses several important open challenges and highlights exciting future research directions in FCL for Edge-AI.
Finally, Section \ref{sec:con} concludes this survey.

\section{Federated Class Continual Learning}
\label{sec:fccl}

The first FCL scenario that we categorized is federated class continual learning (FCCL).
Specifically, 
the objective of class continual learning (CCL) is to discriminate between incrementally observed new classes. 
For instance, a well-trained model in the CCL setting should distinguish all classes, such as `birds' and `dogs' in the first task, `tigers' and `fish' in the second task \cite{zhou2023deep}. 
However, CCL is regarded as a challenging setting since the task identity is not provided. 
Moreover, by integrating into the federated learning paradigm where clients gather data on new classes in a streaming manner, FCCL leads to exacerbated forgetting challenge of old classes during training \cite{dong2022federated, zhang2023target, marfoq2023federatede}. 
Specifically, there are two challenges in FCCL:
\begin{itemize}
    \item \textbf{Challenge 1: intra-task forgetting}: In scenarios where a client is not involved in a particular training round, the newly aggregated global model is at risk of experiencing a performance drop to retain knowledge previously contributed by that client's data. Consequently, this can result in unsatisfying performance when applying the global model to the local data of the non-participating client.
    \item \textbf{Challenge 2: inter-task forgetting}: When clients train models with new tasks, the performance of new global degrades on old tasks.
\end{itemize}

To provide a clear understanding of the FCCL problem, we first formalize its definition as follows.
Given a global server $S_g$ and $M$ clients in the FL process, in each federated training round $r \in \{1, ..., R\}$, each client $m_k (k \in [1, ..., M])$ trains its local model parameters $\theta_{m_k}^r$ by using a sequence of continual tasks $T_{{m_k}} = \{t_{{m_k}}^1, t_{{m_k}}^2, ..., t_{{m_k}}^{t}, ...\}$.
Each task $t_{{m_k}}^t = \{({x^t_i, y^t_i)}\}_{i=1}^{N_{t}}$ consists $N_{t}$ pairs of sample $x^t_i$ and corresponding label $y^t_i$.
The class set $C^t$ of task $t_{{m_k}}^t$ includes its new classes and old class set $C^{t-1}$ in previous $t_{{m_k}}^{t-1}$ tasks.
After local training is complete, each client $m_k$ transmits the updated model parameters $\theta_{m_k}^r$ to the server $S_g$, and server $S_g$ aggregates them into the global parameter $\theta_{G}^r$ to integrate the task knowledge across all clients.
Finally, the server $S_g$ distributes the global parameter $\theta_{G}^r$ to all participating clients in the next training round.

Next, inspired by the tri-level (data-centric, model-centric, and algorithmic) division in CCL \cite{zhou2023deep}, we categorise existing methods into four distinct groups: 
i) \textit{Generative Replay} (Section \ref{subsec-replay}), which falls under the data-centric FCCL; 
ii) \textit{Parameter Regularization} (Section \ref{subsec-regularization}), and iii) \textit{Parameter Decomposition} and \textit{Prompt-based methods} (Section \ref{subsec-decomposition} and \ref{subsec-prompt} respectively), which are the model-centric FCCL; 
and iv) \textit{Knowledge Distillation} (Section \ref{subsec-distillation}), which aligns with the algorithmic FCCL.
This categorization aims to provide a structured overview and facilitate a deeper understanding of the rapidly evolving field. 
Finally, in Section \ref{relation-FCCL}, we summarize these FCCL approaches and analyse the relation between them.

\subsection{Generative Replay}      
\label{subsec-replay}
Replay is a critical strategy to recreate or preserve representations of old classes and combine them with available training data to address the catastrophic forgetting problem in FCCL. 

Shenaj \textit{et al.} \cite{shenaj2023asynchronous} proposed a federated learning system with prototype aggregation for continual representation (FedSpace), which utilized class prototypes in feature space for each old class as replay and contrastive learning to 
preserve previous knowledge to avoid too divergent behaviour between different clients.
Specifically, each client receives the initialized pre-training model over a custom-generated fractal dataset on the server side. 
The client then computes the prototypes of each class, and aggregates them with a weights parameter followed by prototype augmentation.
Furthermore, they introduced a prototype-based loss and an additional loss function based on contrastive learning for clients' optimization.

Hendryx \textit{et al.} \cite{hendryx2021federated} proposed the federated prototypical networks to facilitate more efficient sequential learning of new classes building on the prototypical networks. Specifically, it can enhance model performance through replaying feature vectors representative classes. 
However, it overlooks the non-IID data distribution across these distinct clients. 
Recent works demonstrated that this can be overcome via generative replay.

Generative replay (GR), typically implemented through generative adversarial networks (GANs) \cite{goodfellow2014generativea}, acts as a data replay method by modelling the class distribution of real samples and then synthesizing instances. 
However, adapting GR to FCL settings is not straightforward.
To solve this challenge, Qi \textit{et al.} \cite{qi2022better} discovered empirically that the unstable learning process from distributed training on non-IID data using standard federated learning algorithms can significantly impair GR-based models' performance. 
In response, they proposed FedCIL through model consolidation and consistent enforcement. 
On the server side, the global model is initialized with combined parameters and a collection of classification heads from various clients, then consolidated using instances synthesized by client generators. 
This can avoid failure caused by simply merging the parameters originating from clients with imbalanced new class data. 
To enforce consistency on the client side, a consistency loss is applied to the output logits of the client's classification module during local training. 

Babakniya \textit{et al.} \cite{babakniya2023don, babakniya2023data} introduced the mimicking federated continual learning (MFCL), a method akin to FedCIL, designed to compensate for the lack of old data through generative replay and mitigate forgetting. 
In MFCL, the generative model which encourages to synthesize more uniform and balanced class images is trained on the server side in a data-free manner. 
This enables MFCL to reduce both the local training time and computational costs for clients, and it doesn't necessitate access to their private data.

Recently, considering that exemplar-based methods may not be suitable for privacy-sensitive scenarios, Zhang \textit{et al.} \cite{zhang2023target} proposed TARGET which is an effective solution for addressing catastrophic forgetting in FCCL without storing local private client data or any datasets. 
They first experimentally confirmed that non-IID settings can intensify the catastrophic forgetting problem in FL. 
Then, they used the previously trained global model to transfer knowledge of old tasks to current ones at the model level. 
Additionally, a trained generator synthesizes data to simulate non-IID training datasets with assistant model distillation on the clients at the data level. 
Therefore, TARGET does not require extra datasets or the retention of private data from previous tasks, making it especially suitable for data-sensitive environments.

\subsection{Parameter Regularization}   
\label{subsec-regularization}
Parameter regularization in FCCL is crucial to achieve the dual goals of adapting to new tasks or data distributions while preserving previously acquired knowledge. 
This method evaluates the significance of each network parameter and assigns greater weight to more critical parameters, thus minimizing catastrophic forgetting.

However, it often cannot accumulate comprehensive data on all classes due to the limited storage capacity of FL clients. 
To tackle this challenge, Dong \textit{et al.} \cite{dong2022federated} proposed the global-local forgetting compensation (GLFC) model. 
This work targets local forgetting caused by the class imbalance in local clients by implementing a class-aware gradient compensation loss and a class-semantic relation distillation loss. 
These losses aim to balance the forgetting of old classes and maintain consistent inter-class relations across tasks. 
Moreover, a proxy server is introduced to select the best old global model for aiding each client's local training to address global forgetting caused by the non-IID distribution of classes among clients. 
Further, a gradient-based prototype sample communication mechanism is developed to safeguard the privacy of communications between the proxy server and clients.
Then, Dong \textit{et al.} \cite{dong2023no} extended GLFC and proposed the local-global anti-forgetting (LGA), which surpasses GLFC by efficiently performing local anti-forgetting on old classes. 
They proposed a category-balanced gradient-adaptive compensation loss and a category gradient-induced sematic distillation loss to solve local catastrophic forgetting on old categories.
A proxy server is designed to collect perturbed prototype images of new classes, which can help select the best old model for global anti-forgetting via self-supervised prototype augmentation. 
Compared to the experiments in GLFC, this work conducted more detailed experiments on representative datasets under various FCCL settings and metrics such as top-1 accuracy, F1 score, and recall. 

Apart from these two works, Dong \textit{et al.} \cite{dong2023federated} observed that challenges in federated incremental semantic segmentation (FISS) are heterogeneous forgetting of old classes from both intra-client and inter-client perspectives. 
Therefore, they developed the forgetting-balanced learning (FBL) model to tackle these challenges. 
Specifically, they introduced a forgetting-balanced semantic compensation loss and a forgetting-balanced relation consistency loss to handle intra-client heterogeneous forgetting across old classes, guided by confidently generated pseudo-labels through adaptive class-balanced pseudo-labelling. 
Then, a task transition monitor is designed to surmount inter-client heterogeneous forgetting, enabling new class recognition under privacy protection and storing the latest old model for global relation distillation.

Inspired by the connection between client drift in FL caused by clients' unbalanced classes and catastrophic forgetting for old classes in CL, Legate \textit{et al.} \cite{legate2023re} introduced \textit{local client forgetting} problem. 
Motivated by the balanced softmax cross-entropy method for CL, they applied a re-weighted softmax (WSM) for the loss function of each client based on its class distribution. 
Their method added a regularization term reflecting the class proportions in the client dataset to the standard cross-entropy loss, reducing excessive pressure and subsequent loss on other client data. 

Hu \textit{et al.} \cite{hu2022federated} designed a new FCL framework called DuAFed, featuring a dual attention mechanism for the scenario of different class increments and unbalanced features of clients. 
DuAFed first ensures a balanced pre-training sample distribution by randomly sampling an equal number of instances from each client. 
Further, the iCaRL strategy \cite{rebuffi2017icarl} is employed to accommodate dynamic changes in training tasks. 
To mitigate the noise generated by clients with an imbalanced quantity of classes, a channel attention mechanism is added on the client side, where feature compression, feature map retrieval and regularization via learned weight coefficients of each channel with all the elements of the corresponding channel are successively performed.
Moreover, to solve the challenge that the respective features are unbalanced and the importance is difficult to capture in FCCL, they introduced a feature attention mechanism, which can capture the hierarchical importance of the neural network in multiple local models, for the model aggregation of clients. 

Yao \textit{et al.} \cite{yao2020continual} proposed federated learning with local continual training (FedCL) leveraging a parameter-regularization constrained local continual learning strategy to mitigate the weight divergence and continually integrate knowledge on different local models into the global model, whose efficiency is verified under the different non-IID class data distribution. 
Specifically, they utilized the diagonal of its Fisher information matrix in EWC \cite{kirkpatrick2017overcoming} to evaluate the importance weight matrix of the global model on a small proxy dataset on the server. 
This matrix is used in the loss function to force the local model to fit the local data distribution.

\subsection{Parameter Decomposition}  \label{subsec-decomposition}
Parameter decomposition used in FCCL is a method where the model’s parameters are usually divided into shared global parameters and task-specific parameters, which capture the general knowledge among all learned tasks and informative knowledge for tasks with specific classes. This enables the learned global model to adapt to new class-incremental tasks without losing the previous knowledge learned from previous tasks. 

Motivated by additive parameter decomposition (APD) \cite{yoon2019scalable}, Yoon \textit{et al.} \cite{yoon2021federated} proposed FedWeIT.
They decomposed the model parameters into dense global parameters and sparse task-specific parameters to maximize the knowledge transfer between clients while minimizing the interference of irrelevant knowledge from other clients and communication costs. 
Further, they divided task-specific parameters into local base parameters and task-adaptive parameters, which capture the general knowledge for each client and each task with specific classes per client, respectively. 
Moreover, the sparse mask is applied to select only relevant base parameters for the knowledge of specific classes to lower the number of parameters transferred, thus reducing communication costs.

Similar to FedWeIT, Zhang \textit{et al.} \cite{zhang2022cross} proposed cross-FCL based on parameter decomposition inspired by APD \cite{yoon2019scalable} and several cross-edge strategies, which is a cross-edge FCL algorithm to enable cross-edge devices to continually learn tasks without forgetting. 
They used parameter decomposition by only aggregating base parameters from given tasks with specific classes to solve the challenge of knowledge interference from model aggregation in FL and from inter-task knowledge in CL. 
In addition, to tackle the cross-edge initial decision for usage between the local model and global model, several different cross-edge strategies including discard, replace, finetune, fusion and EWC fusion are proposed for different task relationships.

Different from the above methods, Luo \textit{et al.} \cite{luopan2023fedknow} proposed an FCL framework called FedKNOW, which continually extracts and integrates the knowledge of signature tasks, featuring the concept of signature tasks that are the most dissimilar tasks identified from local past tasks. 
Specifically, each client consists of a knowledge extractor, a gradient restorer and a gradient integrator. 
FedKNOW first retained the top-ranked weight parameters that are extracted as specified tasks’ knowledge and restored the specified number of previous gradients that are most dissimilar tasks with the current task’ gradient to prevent catastrophic forgetting, based on the weight-based pruning technique. 
Then, the gradient integrator is designed to mitigate negative knowledge transfer and improve substant model performance by incorporating gradients from before aggregation and after aggregation. 
It is worth noting that FedKNOW as a client-side solution is more scalable than FedWeIT as a server-side solution for FCCL due to the lower communication cost.

\subsection{Prompt-based methods}   \label{subsec-prompt}

Recent trends involve designing FCCL approaches using a pre-trained Vision Transformer (ViT) as a backbone. 
The ViT adapts its representational capability to streaming class-incremental data through a continual learning method known as `learning to prompt', by dynamically incorporating a set of learned model embeddings, i.e., prompts.

Halbe \textit{et al.} \cite{halbe2023hepco} first formulated \textit{intra-task forgetting} and \textit{inter-task forgetting} in FCCL. 
To mitigate forgetting while minimizing communication costs, protecting client privacy, and enhancing client-level computational efficiency, they proposed HePCo, a prompt-based data-free FCCL method. 
In HePCo, each client performs decomposed prompting, where prompts holding class-specific task information are used to solve corresponding local tasks. 
The final prompt is obtained with a weighted summation by the cosine scores of all these prompts. 
During local learning, each client learns the key and prompt matrices along with the classifier while keeping the ViT backbone frozen. 
Then, a few parameters including key, prompt weights and classifier weights are transferred to the server, which lowers communication overhead and safeguards client privacy by preventing local model inversion. 
On the server side, the latent generator, which takes as input a class label encoded using an embedding layer and a noise vector sampled from the standard normal distribution is trained for the current and previous tasks. 
Once generator training is finished, data-free distillation in the latent space is then employed to combat intra-task forgetting for the current task through finetuning the server model while using pseudo-data corresponding to past tasks helps mitigate inter-task forgetting.

Bagwe \textit{et al.} \cite{bagwe2023fedcprompt} pointed out that the challenge in implementing prompting techniques in FCCL is the unbalanced class distribution of distributed clients, which can cause biased learning performance and slow convergence, while asynchronous task appearances further deteriorate it. 
They proposed Fed-CPrompt by incorporating asynchronous prompting learning and contrastive and continual loss (C2Loss) to alleviate inter-task forgetting and inter-client data heterogeneity.
Fed-CPrompt allows class-specific task prompts aggregation in parallel by taking advantage of task synchronicity. 
C2Loss is designed to accommodate discrepancies due to biased local training between clients in environments with heterogeneous data distribution and to curb the forgetting effect via enforcing distinct task-specific prompts construction.

Differing from previous methods, motivated by the intuition that task-irrelevant prompts may contain potential common knowledge to enhance the embedded features,  
Liu \textit{et al.} \cite{liu2023federated} integrated three types of prompts (i.e., task-specific, task-similar prompts and task-irrelevant prompts) into image feature embedding. 
This strategy effectively preserves both old and new knowledge within local clients, thereby addressing the issue of catastrophic forgetting. 
Additionally, it ensures the thorough integration of knowledge related to the same task across different clients 
via sorted and aligned the task information in the prompt pool. 
This effectively mitigates the non-IID problem, which arises due to class imbalances among various clients engaged in the same incremental task.

\subsection{Knowledge Distillation}     \label{subsec-distillation}
Knowledge Distillation (KD) \cite{hinton2015distillinga}, initially developed for transferring knowledge from larger and complex models to smaller and compact models, has gained widespread use in FCCL. 
It enables an `old model' to aid the currently updating `new model'. 
Although learning without forgetting (LwF) \cite{li2018learninga} was the first successful application of KD in CCL, it can not directly apply to the federated learning framework due to the centralized nature of CCL. 
To address this, 
Usmanova \textit{et al.} \cite{usmanova2022federated, usmanova2021distillation} introduced FLwF to recognize the human activity based on all incrementally seen classes of behaviour from local clients on the 6 classes representing different human activities in the UCI HAR dataset.
FLwF, which first extended KD to the federated setting, is the implementation of a standard LwF method in FCL consisting of the past model of a client as the teacher model and one current client model as the student model.
Additionally, FLwF-2T, which consists of two teacher models including the past model of a client and the server, was proposed to reduce forgetting in FCCL by leveraging a server that maintains a general knowledge base across all clients' class distribution.

Ma \textit{et al.} \cite{ma2022continual} proposed continual federated learning with distillation (CFeD), which performs KD at both client and server levels, uniquely featuring an independent unlabeled surrogate dataset for each client. 
Specifically, it introduces a client division mechanism to utilize under-exploited computational resources, aiding in reducing inter-task forgetting. 
Additionally, inspired by the mini-batch iterative update approach in centralized training, server-side distillation is designed to alleviate intra-task forgetting.
In their class continual learning experimental scenarios, CFeD outperforms other baselines, demonstrating the advantage of using the surrogate dataset to obtain reasonable soft labels for old tasks.

Wei and Li \cite{wei2022knowledge} developed the federated learning with knowledge lock (FedKL) to tackle the issue of catastrophic forgetting in federated learning, particularly the loss of knowledge from other participants due to local updates. 
FedKL utilizes KD techniques to preserve previously acquired knowledge while overcoming server knowledge forgetting caused by data isolation.

Efforts to expand FCCL into areas beyond computer vision, such as intrusion detection \cite{jin2024fl}, are emerging. Jin \textit{et al.} \cite{jin2024fl} introduced FL-IIDS to solve catastrophic forgetting in federated intrusion detection systems (IDS). 
However, this approach simplifies the challenge by assuming that traffic data across local clients in FL-IIDS is IID. 
They identify three key issues in real-world intrusion detection: (1) class imbalance in various traffic data types, (2) a predominance of new over old classes in current tasks, leading to a bias towards new knowledge, and (3) a shrinking in the sample size of old classes in dynamic example memory, weakening the ability to learn old classes. 
To combat these, they proposed dynamic example memory, class gradient balancing loss, and sampling label smoothing loss, respectively. 
Notably, their KD strategy, termed label smoothing loss, incorporates soft labels of old classes into current training, enhancing the model's generalization over old classes and mitigating local model forgetting.

\begin{table}[htbp]
\centering
\caption{Major Contribution of FCCL Methods}
\label{tbl-FCCL}  
\begin{tabular}{@{}p{2.0cm} p{1.0cm} p{10cm}@{}}
    \toprule
   \multicolumn{1}{l}{\textbf{Approach}} & \multicolumn{1}{c}{\textbf{Paper}} & \multicolumn{1}{c}{\textbf{Key Contribution}} \\
    \midrule
    \multicolumn{1}{c}{ \multirow{5}{2.0cm}{Generative Replay (Section \ref{subsec-replay})} }   
    & \multicolumn{1}{c}{\multirow{1}{*}{ 
    \cite{shenaj2023asynchronous} }} &  Asynchronous FCL with class prototypes replay  \\ \cline{2-3}     
    & \multicolumn{1}{c}{\multirow{1}{*}{ 
    \cite{hendryx2021federated} }} &  Federated prototypical networks  \\ \cline{2-3}     
    & \multicolumn{1}{c}{\multirow{1}{*}{ 
    \cite{qi2022better} }} & Model consolidation and consistent enforcement  \\ \cline{2-3} 
    & \multicolumn{1}{c}{\multirow{1}{*}{ 
    \cite{babakniya2023don,babakniya2023data} }} & Compensate for the absence of old data by a data-free generative replay \\ \cline{2-3} 
    & \multicolumn{1}{c}{\multirow{1}{*}{ 
    \cite{zhang2023target} }} & GR and KD within an exemplar-free continual learning  \\ \hline

    \multicolumn{1}{c}{ \multirow{9}{2.0cm}{Parameter Regularization (Section \ref{subsec-regularization})} } 
    &\multicolumn{1}{c}{\multirow{1}{*}{
    \cite{dong2022federated} }}  & The first work to alleviate local and global forgetting in FCCL \\ \cline{2-3} 
    & \multicolumn{1}{c}{\multirow{2}{*}{ 
    \cite{dong2023no} }} & A category-balanced gradient-adaptive compensation loss and a category gradient-induced semantic distillation loss \\ \cline{2-3} 
    & \multicolumn{1}{c}{\multirow{1}{*}{  
    \cite{dong2023federated} }} & The first global continual segmentation model for FISS \\ \cline{2-3} 
    & \multicolumn{1}{c}{\multirow{1}{*}{ 
    \cite{legate2023re} }} & Re-weight the softmax logits prior to computing the loss \\  \cline{2-3} 
    & \multicolumn{1}{c}{\multirow{2}{*}{ 
    \cite{hu2022federated} }} & Channel attention NN model and federated aggregation algorithm based on the feature attention mechanism \\ \cline{2-3} 
    & \multicolumn{1}{c}{\multirow{1}{*}{ 
    \cite{yao2020continual} }} & Importance weight matrix for better initialization of federated models \\ \hline

    \multicolumn{1}{c}{ \multirow{5}{2.0cm}{Parameter Decomposition (Section \ref{subsec-decomposition})}  }
    & \multicolumn{1}{c}{\multirow{1}{*}{
    \cite{yoon2021federated} }} & Weighted inter-client transfer based on task-specific parameters  \\ \cline{2-3} 
    & \multicolumn{1}{c}{\multirow{2}{*}{
    \cite{zhang2022cross} }} & Task-specific parameter aggregation and cross-edge strategies for initial decision for federated models \\ \cline{2-3} 
    & \multicolumn{1}{c}{\multirow{2}{*}{
    \cite{luopan2023fedknow} }} & knowledge extraction and gradient restoration based on weight-based pruning, and gradient integration  \\ \hline

    \multicolumn{1}{c}{ \multirow{5}{2.0cm}{Prompt-based methods (Section \ref{subsec-prompt})}  }
    & \multicolumn{1}{c}{\multirow{2}{*}{
    \cite{halbe2023hepco} }} & A lightweight generation and distillation scheme to consolidate client models at the server based on prompting \\ \cline{2-3} 
    & \multicolumn{1}{c}{\multirow{1}{*}{
    \cite{bagwe2023fedcprompt} }} & Asynchronous prompt learning and contrastive continual loss \\ \cline{2-3} 
    & \multicolumn{1}{c}{\multirow{2}{*}{
    \cite{liu2023federated} }} & A rehearsal-free FCL method based on prompting with the consideration of privacy and limited memory  \\ \hline
    
    \multicolumn{1}{c}{ \multirow{6}{2.0cm}{Knowledge Distillation (Section \ref{subsec-distillation})} }
    & \multicolumn{1}{c}{\multirow{1}{*}{ 
    \cite{usmanova2022federated, usmanova2021distillation} }} & The first work to extend LwF to the federated setting\\   \cline{2-3} 
    & \multicolumn{1}{c}{\multirow{2}{*}{ 
    \cite{ma2022continual} }} & A client division mechanism and the server distillation with the unlabeled surrogate dataset \\  \cline{2-3} 
    & \multicolumn{1}{c}{\multirow{1}{*}{ 
    \cite{wei2022knowledge} }} & Overcoming the server knowledge forgetting caused by data isolation \\  \cline{2-3} 
    & \multicolumn{1}{c}{\multirow{2}{*}{ 
    \cite{jin2024fl} }} & Sample label smoothing loss function leveraging KD to enhance the local model memory \\
    \bottomrule
  \end{tabular}
\end{table}

\begin{figure}[htbp]
    \centering
    \includegraphics[scale=0.6]{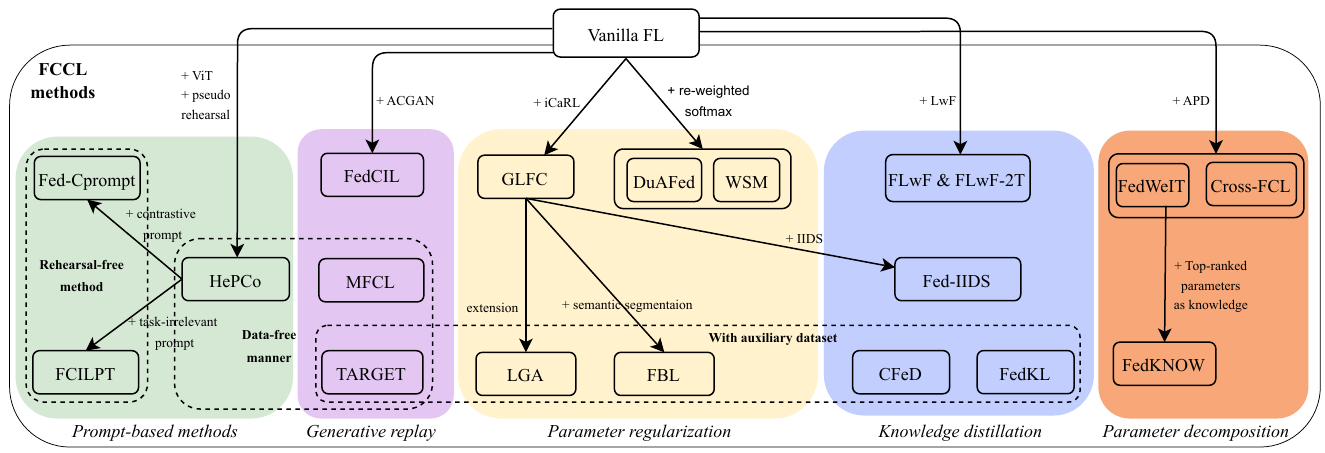}
    \caption{Diagram of the relation among FCCL methods. There are five categories in our paper: generative replay ({\color{purple}purple}), parameter regularization ({\color{ACMYellow}yellow}), parameter decomposition ({\color{orange}orange}), prompt-based methods ({\color{green}green}) and knowledge distillation ({\color{blue}blue}). Auxiliary datasets, the data-free manner with distillation and rehearsal-free methods, frequently employed in some methods, are indicated by the dashed boxes.}
    \label{fig: relation-FCCL}
\end{figure}

\subsection{Summary and analysis of FCCL approaches} \label{relation-FCCL}
The major contribution of FCCL methods categorised above is summarised in Table \ref{tbl-FCCL}. 
To further indicate the relation between these representative methods,
Fig. \ref{fig: relation-FCCL} distinguishes these methods using generative replay ({\color{purple}purple}), parameter regularization ({\color{ACMYellow}yellow}), parameter decomposition ({\color{orange}orange}), prompt-based methods ({\color{green}green}) and knowledge distillation ({\color{blue}blue}). 
Moreover, as indicated by the dashed boxes, three common strategies and methods, i.e., auxiliary datasets, the data-free manner with distillation, and the rehearsal-free method, are utilized in several FCCL methods. 
From this diagram, we can see that:
\begin{itemize}
    \item The impact of GLFC \cite{dong2022federated} on subsequent research and the integration of techniques from different fields are evident.
    \item The impact of LwF \cite{li2018learninga}, iCaRL \cite{rebuffi2017icarl} and APD \cite{yoon2019scalable} upon the field of FCCL is huge.
    \item With the rapid development of generative models, techniques represented by GAN have emerged in the field of replay-based FCCL, which especially satisfies the users' data privacy protection needs in a data-free manner.
    \item Benefiting from the advanced capabilities and rapid developments of foundation models in representation and transferability, a notable emergence of innovative FCL methods incorporating well-pre-trained models, such as ViT-based methods, have progressively surfaced \cite{halbe2023hepco,liu2023federated,liu2023fedet}.
    
\end{itemize}
These approaches offer various solutions and pivotal insights to address the challenges encountered in FCCL which is still in its nascent stage.

\section{Federated Domain Continual Learning}
\label{sec:fdcl}
In this section, we delve into various strategies aimed at addressing challenges in the second FCL scenario, namely Federated Domain Continual Learning (FDCL). 
In FDCL, ‘domain’ typically refers to the distribution of datasets. 
Traditional continual learning focuses on the dynamics of individual domains, while the integration of FL further allows each client with its private dataset to be treated as a separate domain. 
Therefore, FDCL research focuses on the generalization of different domains and the dynamic adaptation of individual domains.

To provide a clear understanding of the problem definition of FDCL, we first conceptualize it as follows.
For a given period \( [0, T] \) and $K$ clients engaged in FL, we assume that the \( k\)th client contains a set of sample $x$ and label $y$ pairs \( D^t_k = \{(x^t_i, y^t_i)\}_{i=1}^{|K|} \) at a given time \( t\). 
It is worth noting that there may be unlabeled samples in some clients, but all samples belong to known classes. 
Multiple clients in FL form a global known domain \( D^t_g = \{D^t_1, D^t_2, \ldots, D^t_K\} \). 
Subsequently, local training and aggregation are performed in the classical FL paradigm. 
In this way, each client constructs a local model \( f_{\theta_k}: X_k \rightarrow Y_k \) with its private dataset, which is aggregated by the server to generate a comprehensive global model \( f_{\theta_g}: X_g \rightarrow Y_g \) after multiple rounds of communication. 
In FDCL scenarios, the global model \(  f_{\theta_g} \) not only serves as a representation of the known domain \( D^t_g \) but also can be used to generalize the unknown domain \( D^t_{unk} \). Furthermore, the local domain changes over time, which means \( D^t_k \neq D^{t+d}_k \) after a time interval  \( d \). 
Moreover, in FDCL, the task identity is not necessary during testing, because if each task has the same classes, the output would be the same as well.

\begin{figure}[htbp]
\centering
\includegraphics[width=0.7\linewidth]{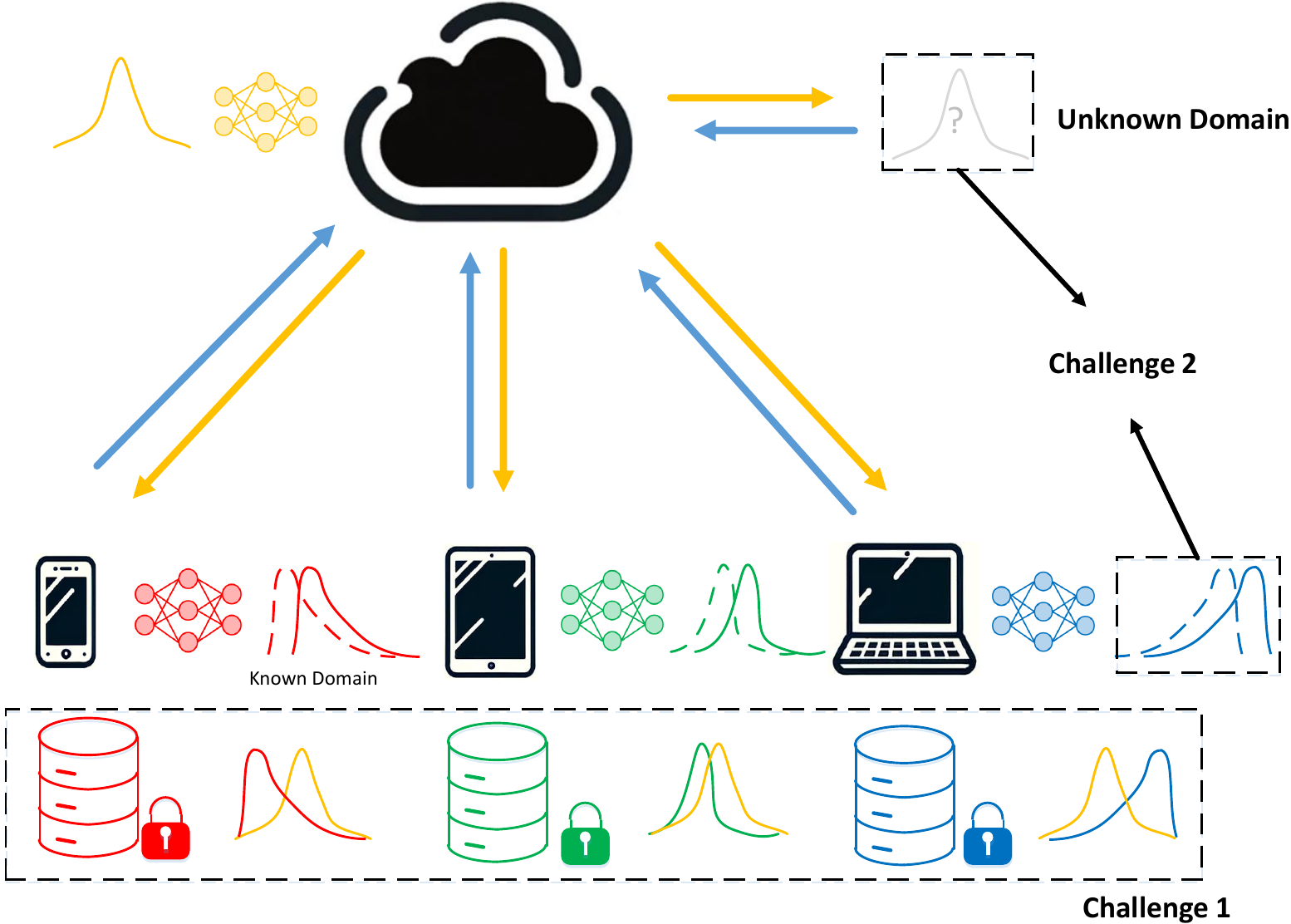}
\caption{Overall diagram of challenges faced by FDCL, challenge 1: privacy protection of multi-source domains and domain drift of the global model concerning the local model (inter-domain); challenge 2: generalization of the global model for the unknown domain, and domain drift of locally known domains over time and data (intra-domain). }
\label{fig:fdcl}
\end{figure}

As illustrated in Fig. \ref{fig:fdcl}, FDCL faces two unique challenges:
\begin{itemize}
\item \textbf{Challenge 1: distributed multi-source domains generalization based on privacy protection}: In the distributed environment of FL, each client constitutes a separate domain. This diversity significantly increases the challenges associated with the global model generalization. Moreover, the commitment to protecting privacy results in data isolation, which intensifies the complexity of learning and optimizing the global model.

    \item \textbf{Challenge 2: unknown domain generalization and known domain drift}: In the context of FDCL, the global model also needs to extend its generalization capabilities beyond the multi-source domains to cover unknown domains. Furthermore, dynamic data changes in continual learning result in known domain drift, so models need to have the ability to learn and adapt efficiently in a time-evolving and uncertain data environment.
\end{itemize}

Overall, these two challenges outline the balance between maintaining data privacy and improving model generalization capabilities in a changing environment. 
According to the different types of approaches, current studies on FDCL can be divided into four main areas: 
\textit{Domain Data Supplementation} (Section \ref{sec:domain_data_supplementation}), \textit{Domain Knowledge Learning} (Section \ref{sec:domain_knowledge_learning}), \textit{Domain Model Enhancement} (Section \ref{sec:domain_model_enhancement}), and \textit{Domain Weight Aggregation} (Section \ref{sec:domain_weight_aggregation}). 
Various research approaches and the key contributions of FDCL are detailed in Table \ref{tab:fdcl}.  
In the subsequent subsections, we thoroughly examine the approaches associated with these four areas.

\begin{table}[htbp]
  \centering
  \caption{Different Research Approaches in FDCL}
  \begin{tabular}{@{}p{2.5cm} p{2cm} p{8cm}@{}}
    \toprule
   \multicolumn{1}{c}{\textbf{Type of Approaches}} & \multicolumn{1}{c}{\textbf{Paper}} & \multicolumn{1}{c}{\textbf{Key Contribution}} \\
    \midrule
    \multicolumn{1}{c}{ \multirow{6}{2.5cm}{Domain Data Supplementation (Section \ref{sec:domain_data_supplementation})} }    
    & \multicolumn{1}{c}{\multirow{1}{*}{ \cite{liu2021feddg} }} & Continuous frequency space interpolation \\ \cline{2-3} 
    & \multicolumn{1}{c}{\multirow{2}{*}{ \cite{liu2022overcoming} }} & Training generators on the central server for clients to produce synthetic global data \\ \cline{2-3} 
    & \multicolumn{1}{c}{\multirow{1}{*}{ \cite{park2021tackling} }} & Variable embedding rehearsal and server-side training  \\ \cline{2-3} 

    &\multicolumn{1}{c}{\multirow{1}{*}{ \cite{casado2020federated, casado2022concept,casado2023ensemble} }}  & Client domain drift detection and adaptation \\ \cline{2-3} 
    & \multicolumn{1}{c}{\multirow{1}{*}{ \cite{zhang2023spatial} }} & Storing generalized representations of local data \\ \hline
    
    \multicolumn{1}{c}{ \multirow{7}{2.5cm}{Domain Knowledge Learning (Section \ref{sec:domain_knowledge_learning})} } 
    & \multicolumn{1}{c}{\multirow{2}{*}{ \cite{huang2022learn} }} & Federated cross-correlation learning and dual-domain knowledge distillation loss \\ \cline{2-3} 
    & \multicolumn{1}{c}{\multirow{1}{*}{ \cite{ma2022continual} }} & Continuous distillation federated learning \\ \cline{2-3} 
    & \multicolumn{1}{c}{\multirow{1}{*}{ \cite{wang2023multi} }} & Lifetime federated meta-reinforcement learning \\  \cline{2-3} 
    & \multicolumn{1}{c}{\multirow{1}{*}{ \cite{guo2021towards} }} & Incremental unsupervised adversarial domain adaptation \\ \cline{2-3} 
    & \multicolumn{1}{c}{\multirow{1}{*}{ \cite{huang2022incremental} }} & Client-side and time-drift modeling \\ \cline{2-3} 
    & \multicolumn{1}{c}{\multirow{1}{*}{ \cite{chen2023learning} }} & Weights dynamic update \\ \hline

    \multicolumn{1}{c}{ \multirow{11}{2.5cm}{Domain Model Enhancement (Section \ref{sec:domain_model_enhancement})}  }
    & \multicolumn{1}{c}{\multirow{1}{*}{ \cite{de2023continual} }} & Echo state networks and intrinsic plasticity \\ \cline{2-3} 
    & \multicolumn{1}{c}{\multirow{1}{*}{ \cite{zhang2023communication} }} & Synaptic intelligence in FCL \\ \cline{2-3} 
    & \multicolumn{1}{c}{\multirow{1}{*}{ 
    \cite{chathoth2022differentially} }} & Differential privacy and synaptic intelligence \\ \cline{2-3} 
    & \multicolumn{1}{c}{\multirow{1}{*}{ 
    \cite{ma2022efl} }} & Elastic federated learning \\ \cline{2-3} 
    & \multicolumn{1}{c}{\multirow{1}{*}{ \cite{bereska2022continual} }} & Reservoir computing and recurrent neural network\\   \cline{2-3} 
    & \multicolumn{1}{c}{\multirow{1}{*}{ \cite{mori2022continual} }} & Progressive neural network \\   \cline{2-3} 
    & \multicolumn{1}{c}{\multirow{1}{*}{ \cite{le2021federated} }} & Broad learning in FCL \\ \cline{2-3} 
    & \multicolumn{1}{c}{\multirow{2}{*}{ \cite{zhu2022attention} }} & Radial basis function and self-organizing incremental neural network \\  \cline{2-3} 
    
    & \multicolumn{1}{c}{\multirow{1}{*}{ \cite{han2022lightweight} }} & Very fast decision tree and order-preserving encoding \\ \hline

    \multicolumn{1}{c}{ \multirow{7}{2.5cm}{Domain Weight Aggregation (Section \ref{sec:domain_weight_aggregation})} }
    & \multicolumn{1}{c}{\multirow{1}{*}{ \cite{zhang2023federated} }} & Genetic algorithm and domain flatness constraint\\   \cline{2-3} 
    & \multicolumn{1}{c}{\multirow{2}{*}{ \cite{dupuy2022learnings} }} & Non-uniform device selection for natural language understanding \\  \cline{2-3} 
    & \multicolumn{1}{c}{\multirow{1}{*}{ \cite{wang2022secure} }} & Orthogonal gradient aggregation \\  \cline{2-3} 
    & \multicolumn{1}{c}{\multirow{1}{*}{ \cite{mawuli2023semi} }} & Semi-supervised FL on evolving data streams \\  \cline{2-3} 
    & \multicolumn{1}{c}{\multirow{1}{*}{ \cite{yao2023finding} }} & Graph-aided FL approach with a few-shot node inhibition \\
    \bottomrule
  \end{tabular}
  \label{tab:fdcl}
\end{table}

\subsection{Domain Data Supplementation}
\label{sec:domain_data_supplementation}
The approach of domain data supplementation aims to achieve local dataset expansion by incorporating the data distribution from other clients. 
It can address the issue of poor generalization caused by dataset isolation from different clients and the absence of old data from the local client. 
A viable strategy involves employing data synthesis techniques to create proxy datasets that mimic others’ domains by \textbf{supplementing data from other clients} while maintaining privacy \cite{liu2021feddg, liu2022overcoming, park2021tackling}. Additionally, direct storage of old data also serves as an effective way as it \textbf{supplements old data from the local client} to remember the previous domain \cite{ casado2020federated, casado2022concept, casado2023ensemble, zhang2023spatial}.

\textbf{Supplementing data from other clients}.
Inspired by feature extraction in image frequency domain space, Liu \textit{et al.} \cite{liu2021feddg} proposed federated domain generalization (FedDG), which exchanges part of frequency information across clients to supplement data in a privacy-conscious manner. 
Specifically, FedDG decomposes the amplitude (i.e., low-level distribution) and phase signals (i.e., high-level semantics) by fast fourier transform (FFT). 
Based on this, an ‘amplitude distribution bank’ is created for client data sharing, where each client generates new signals by interpolating the local amplitude with data from the shared bank while maintaining the local phase signal constant. 
These interpolated signals are transformed by the inverse fourier transform (IFT) to create a proxy dataset as complements.
FedDG implicitly synthesizes data from other clients, bridging the gap between local and global models.

Instead of synthesizing data locally, Liu \textit{et al.} \cite{liu2022overcoming} trained a data generator on the centre server. 
The server initially collects data from each client as a constant reference point to train the data generator. 
Then, the trained generator and global model are broadcast to each client to produce synthetic data. 
Furthermore, a mechanism of variable weights is also introduced to alleviate the imbalance in the number of local classes across various clients. 
Although the above method of supplementing data from other clients can improve the generalization ability of the global model, there is still some risk of privacy leakage.

To further improve the security of proxy datasets, Park \textit{et al.} \cite{park2021tackling} introduced variable embedding rehearsal (VER) and server-side training (SST) strategies. 
On one hand, the authors used the VER method that combines the security advantages of variable autoencoder (VAE) and embedding-based reformulation (EBR) by generating random representations of a subset of data from each client. 
On the other hand, the SST strategy facilitates training by rehearsing the data representations that have been safely collected from each client avoiding direct access to the original dataset. 

\textbf{Supplementing old data from local client}.
To address the issue of individual domain drift over time in FDCL, Casado \textit{et al.} \cite{ casado2020federated, casado2022concept, casado2023ensemble} utilized different methods to detect and adapt to local client domain drift. 
For domain drift detection, the authors used a CUSUM-type (Cumulative Sum) method based on a beta distribution. 
Building upon the original method, they proposed a sliding window technique to detect changes in the confidence distribution of local classifiers. 
For domain drift adaptation, they gathered data from new domains to update the long-term storage of the local client, thereby preserving the memory of the previous domain. 
Over time, this approach requires adding more long-term storage memory locally, otherwise forgetting will still occur, which may not be applicable in some resource-constrained scenarios.

Since it is impractical to store all the data before the local client domain drift, Zhang \textit{et al.} \cite{zhang2023spatial} proposed a method to periodically store a generalized representation of the local client data, while taking advantage of dynamically changing data to update models with new domain knowledge. 
In addition, the central server also merges new domains from different clients based on the relevance of the spatial and temporal dimensions.

\subsection{Domain Knowledge Learning}
\label{sec:domain_knowledge_learning}
The approach of domain knowledge learning considers the potential privacy risks associated with proxy datasets. 
It prioritizes the direct knowledge transfer over the data transfer approach to address inter-domain differences among different clients and intra-domain drift within the local client. 
In the case of \textbf{inter-domain knowledge learning}, the strategy focuses on efficiently transferring knowledge from other clients to the local client \cite{huang2022learn, ma2022continual, wang2023multi}. 
In the case of \textbf{intra-domain knowledge learning}, it emphasizes the application of previously acquired knowledge to adapt the local domain drift \cite{guo2021towards, huang2022incremental, chen2023learning}.

\textbf{Inter-domain knowledge learning}.
To solve heterogeneity and catastrophic forgetting problems in distributed domains, Huang \textit{et al.} \cite{huang2022learn} presented a method based on federated cross-correlation and continual learning. 
The authors built a cross-correlation matrix across different clients with an unlabeled public dataset and exploited knowledge distillation techniques in the local updating process.  
A new loss function is proposed by federated cross-correlation learning that boosts model similarity while accounting for model diversity. 
The generated loss function is combined with a dual-domain knowledge distillation-based loss function, where the latest model is computed from a mixture of the learned global model and the previous local model. 

Similarly using the knowledge distillation method, Ma \textit{et al.} \cite{ma2022continual} implemented continual federated learning with distillation (CFeD) at both the client and server side with different learning objectives for different clients. 
The essence of CFeD lies in using the learned model in the previous domain to predict a proxy dataset and then utilising the predictions as pseudo-labels to retain knowledge in currently inaccessible domains.
Moreover, CFeD also presents a server distillation mechanism specifically designed to address within-task forgetting in different domains. 
It involves adjusting the aggregated global model to imitate the output from both the previous global domain and the current local domain model. 

Different from the above methods, Wang \textit{et al.} \cite{wang2023multi} developed a complex lifetime federated meta-reinforcement learning (LFMRL) algorithm, which leveraged prior obtained knowledge from federated meta-learning to quickly adapt to new domains. 
Specifically, LFMRL devises a knowledge fusion algorithm that integrates federated meta-learning and dual-attention deep reinforcement learning to update local gradient data and generate shared models. 
Additionally, LFMRL also designs an efficient knowledge transfer mechanism for the rapid learning of new domains in new environments. 
This approach enhances the generalization of the model not only in the known domains but also in the unknown domains.

\textbf{Intra-domain knowledge learning}.
To cope with the problem of domain drift in newly collected data, Guo \textit{et al.} \cite{guo2021towards} proposed incremental unsupervised adversarial domain adaptation (IUADA) that merges FL and adversarial learning. 
This method aims to transfer knowledge from the local target domain to the model learned from labelled data. 
In particular, the local target feature extractor and discriminator are alternately trained through adversarial learning, separating source features into positive and negative. 
Meanwhile, the gradient of the prediction score serves as an attention weight to obtain distinctive features, which are aligned with local domain features to adapt domain drift. 
This adversarial learning approach increases computational complexity and resource requirements, especially for local clients with limited resources.

Regarding local domain drift in dynamic environments, Huang \textit{et al.} \cite{huang2022incremental} constructed a model for local client domain drift evolving. 
The theoretical demonstration in this work reveals that the convergence rate of the method in time-evolving scenarios is related to the approximation accuracy. 
Moreover, Chen and Xu \cite{chen2023learning} introduced a dynamic update mechanism that leverages new weights to adjust the parameters of the output classifier. 
This mechanism allows the model to seamlessly integrate information from recently acquired data while preserving previously learned knowledge.

\subsection{Domain Model Enhancement}
\label{sec:domain_model_enhancement}
The approach of domain model enhancement emphasizes improving the models of local clients to mitigate the issue of memory loss. 
It enhances the model's capabilities of resistance to forgetting and adaptability by integrating innovative network structures or techniques. 
According to whether the structure of the model is fixed, existing works based on this approach can be divided into two categories: \textbf{fixed structure} \cite{ de2023continual, zhang2023communication, chathoth2022differentially, ma2022efl} and \textbf{non-fixed structure} \cite{bereska2022continual, mori2022continual, le2021federated, zhu2022attention, han2022lightweight}.
This approach aims to leverage the learning capabilities of the client model to retain the memory of the old domain while seamlessly integrating new domain knowledge.

\textbf{Fixed Structure}. 
De Caro \textit{et al.} \cite{de2023continual} facilitated effective learning in dynamic environments by employing echo state networks (ESNs) and intrinsic plasticity (IP). 
The authors introduced the FedIP algorithm to optimize the processing of stationary data in a federated learning setting and adapt the learning rules of IP. 
In non-stationary scenarios, FedCLIP extends FedIP by updating memory buffers and sampling the mini-batches. 
Zhang \textit{et al.} \cite{zhang2023communication} exploited synaptic intelligence (SI) for weight updating to maintain the memory of the previous domain. 
They also added a structural regularization loss term that integrates knowledge from other local models to tune the global model towards a global optimum while minimizing weight variance. 
Chathoth \textit{et al.} \cite{chathoth2022differentially} observed that non-IID data distributions have a significant impact on the performance of differential privacy (DP) stochastic algorithms. To counteract this issue, they integrated DP with SI to meet the privacy requirements of each client. In particular, they mitigated catastrophic forgetting by adding a quadratic SI loss to the objective function to minimize modifications to parameters that affect the previous model. Furthermore, they improved the \((\varepsilon, \delta)\)-DP training method for a cohort-based DP setting, tailoring it to meet the distinct privacy requirements of each cohort.
To bridge continual learning with federated learning and improve the robustness of different clients to non-IID problems, Ma \textit{et al.} \cite{ma2022efl} introduced the elastic federated learning (EFL) framework. It integrates an elasticity term that constrains the volatility of crucial parameters, as determined by the Fisher information matrix, within the local objective function. Moreover, they employed scaling aggregation coefficients to counteract convergence degradation. The framework is further optimized through sparsification and quantization techniques, effectively compressing both upstream and downstream communications.

\textbf{Non-fixed Structure}.
Drawing on insights from predictive coding in neuroscience, where updating the parameters of only some of the active heads can prevent the other inactive heads from being forgotten, Bereska \textit{et al.} \cite{bereska2022continual} proposed an approach based on reservoir computing in FCL, which is a state-of-the-art method for training recurrent neural networks (RNN) in dynamic environments. They slowed down weight forgetting by fixing the weights of hidden layers in the RNN and training multiple competing prediction heads simultaneously.
Mori \textit{et al.} \cite{mori2022continual} split the neural network for each client into a unique feature extraction component and a common feature extraction component. 
The authors regarded the local training as learning a unique task without forgetting the knowledge of a common task, thus introducing the progressive neural network (PNN) as the continual learning method in their solution.
Le \textit{et al.} \cite{le2021federated} mitigated the catastrophic forgetting and adapted to environmental changes by broad learning (BL), which supports CL without retraining each client for new data. 
Moreover, they designed a weighted processing strategy and a batch-asynchronous technique to support accurate and fast training. 
This asynchronous update method combined with BL can decouple local training from the knowledge of the global model. 
Zhu \textit{et al.} \cite{zhu2022attention} proposed the SOINN-RBF method, which effectively combines radial basis function (RBF) networks and self-organizing incremental neural networks (SOINN). 
This method aims to optimize data labelling management and real-time sample domain adaptation through high dimensional spatial mapping to improve data regularity identification and generalization. 
Han \textit{et al.} \cite{han2022lightweight} presented an incremental tree model construction method based on very fast decision tree (VFDT) for efficiently handling domain drift. 
They developed a lightweight practically order-preserving encoding (POPE) method, which replaces complex encryption algorithms while reducing computational and communication burdens. 
Additionally, they adapted a region-counting method to effectively reduce the memory overhead of POPE. 

\subsection{Domain Weight Aggregation}
\label{sec:domain_weight_aggregation}
The approach of domain weight aggregation assesses and reorganizes the relationships between clients based on the uploaded model weights. 
Researches based on this approach can be divided into two key parts, \textbf{non-uniform weight aggregation} \cite{zhang2023federated, dupuy2022learnings, wang2022secure} and \textbf{reorganization relationships of clients} \cite{ mawuli2023semi, yao2023finding}. 
The first part concentrates on directly optimizing the weight aggregation process to improve the model’s generalization capabilities, while the second one emphasizes indirectly influencing the weight aggregation process by organizing relationships between clients and further considers adaptation to dynamically changing environments.

\textbf{Non-uniform weight aggregation}
Zhang \textit{et al.} \cite{zhang2023federated} devised an FL-friendly generalization adjustment (GA) method that combines a genetic algorithm with domain flatness constraint to determine the best weights for each client. 
Specifically, the flatness of each domain is evaluated by the difference in generalization between the global and local models.
Meanwhile, domain weights are dynamically adjusted during server aggregation. 
Dupuy \textit{et al.} \cite{dupuy2022learnings} showed that in natural language understanding (NLU) training, non-uniform device selection based on the number of interactions improves model performance, with benefits increasing over time.
Wang \textit{et al.} \cite{wang2022secure} introduced the orthogonal gradient aggregation (OGA) method instead of uniform weight aggregation, which updates gradients orthogonally to prior parameter spaces to prevent catastrophic forgetting in domain transfer, thereby retaining old knowledge and enhancing privacy. 
Although this approach solves the problem of generalization of data from different source domains (known domains), the generalization performance for unknown domains and domain drift problems still needs to be discussed thoroughly.

\textbf{Reorganization relationships of clients}.
Mawuli \textit{et al.} \cite{mawuli2023semi} proposed a semi-supervised federated learning approach on evolving data streams (SFLEDS) that addresses domain drift and privacy protection. 
Their proposed method utilized a distributed prototype-based technique that uses k-means clustering to group data stream instances into micro-clusters.
Then, an error-driven technique is employed to capture inter-and-intra domain drift.
Specifically, it efficiently performs collaborative semi-supervised prediction tasks by merging global and local models and incorporates probabilistic client-server consistency techniques to address domain drift. 

Yao \textit{et al.} \cite{yao2023finding} introduced a graph-aided federated learning (GAFL) approach with a few-shot node inhibition mechanism to improve the generalization capability of global models. 
GAFL designs collaborative graphs of pair-wise and category-wise levels to describe the relationship of customers to distinguish different data distributions. 
A continual learning approach is tailored to new clients, limiting graphics and model updates to a smaller scope, thus minimizing the disruption caused by the original model domain.

\subsection{Summary and analysis of FDCL approaches}
This section summarizes and analyzes the four types of FDCL approaches mentioned above, addressing domain generalization and domain drift adaptation from a unique perspective. 
In summary:

\begin{itemize}
\item \textbf{Domain Data Supplementation}. 
This approach emphasizes strengthening the data-level complements to mitigate data scarcity and diversity issues. 
However, the use of synthetic datasets may pose a risk of privacy leakage, and there is the possibility of using synthetic data to infer the distribution of the original data.
   
\item \textbf{Domain Knowledge Learning}. 
This approach focuses on leveraging acquired knowledge for better application across different clients and drift issues.
Despite the benefits, the complexity of distillation methods may lead to additional computational and communication costs, which may be detrimental to the implementation of some resource-constrained edge devices.
   
\item \textbf{Domain Model Enhancement}. 
This approach aims at improving the model's resistance to forgetting. Specific model enhancement methods may be effective in small datasets or simple tasks, while their generalizability and adaptability to other complex tasks are inconclusive.

\item \textbf{Domain Weight Aggregation}.  
This approach prioritizes the optimization of weight relationships between clients. 
It proves beneficial for known domains, however, its performance in generalizing to unknown domains deserves further investigation.
\end{itemize}

Each approach contributes valuable insights for overcoming challenges in FDCL. 
Nonetheless, they are also accompanied by specific limitations that require further attention.

\section{Federated Task Continual Learning}
\label{sec:ftcl}
In this section, we will investigate the emerging challenges and state-of-the-art research of the third FCL scenario, namely Federated Task Continual Learning (FTCL).

In this scenario, the local clients learn a set of distinct tasks over time for which the task identity is explicitly provided, i.e., the learning algorithm is clear about which task will be executed.
Then global aggregation is performed in multiple rounds to generate a global model enabling the distribution and update of the knowledge for different tasks across clients.

Similar to the previous sections, we first provide a clear formalisation of the FTCL problem.
Considering a global server $S$ and $C$ distributed clients in a federated framework,
each client $c_i \in \{c_1 , ..., c_C \}$ learns a local model on its private task dataset $TD_{c_i}$ with task sequence $\{1, ..., t, ..., T \}$, where $TD_{c_i}^t = \{x^t_j,y^t_j\}^{N^t}_{j=1}$ is a labeled dataset for task $t$ with ${N^t}$ instances of $x^t_j$ and its label $y^t_j$.
In FTCL, there is no relationship among the datasets $TD_{c_i}$ across all clients. 
In each federated training round $r \in \{1, ..., R\}$, each client $c_i$ updates its model parameters $\theta_{c_i}^r$ by using task dataset $TD_{c_i}$ in a task continual learning setting and accelerate the current task learning with learned knowledge from the past tasks.
Then, each client $c_i$ transmits updated model parameters $\theta_{c_i}^r$ after training to the server $S$, and the server $S$ aggregates them into the global parameter $\theta_{G}^r$ to integrate the task knowledge across all clients.
Finally, the server $S$ distributes the global parameter $\theta_{G}^r$ to all participating clients in the next training round.
In FTCL, the task identity is clearly provided during learning and testing, so the model can be trained and performed by referring to the task-specific components.

Under this FTCL scenario, there are two main challenges that need to be solved after each client updates its local model with the global parameter $\theta_{G}$ to obtain the cross-client task knowledge:
\begin{itemize}
    \item \textbf{Challenge 1:} Catastrophic forgetting happens due to the insufficient training data of tasks from other clients in $TD_{c_i}$.
    \item \textbf{Challenge 2:} The performance of local clients degrades as client and task heterogeneity increases, causing local model training to update its parameters in the wrong direction.
\end{itemize}

\begin{figure}[!ht]
  \centering
  \includegraphics[width=1\linewidth]{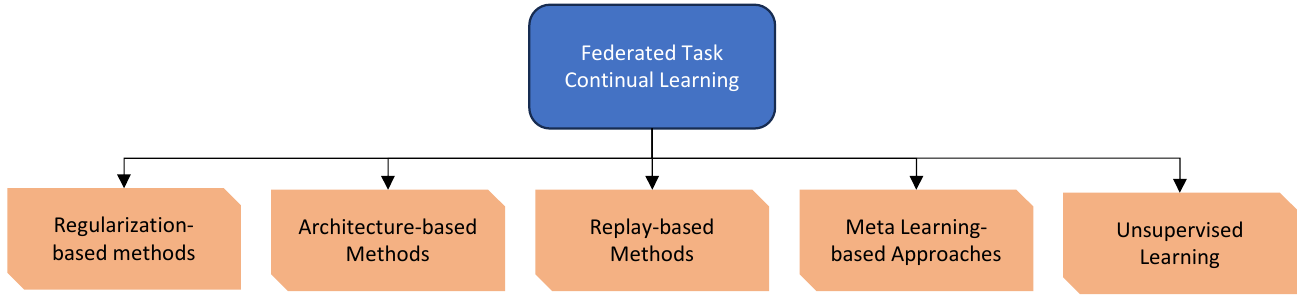}
  \caption{The elaborated taxonomy of representative federated task continual learning methods}
  \label{fig:ftcl}
\end{figure}

Recently, many studies have been conducted to provide different methods to solve these challenges in FTCL.
In the following subsections, we will present an elaborated taxonomy of representative federated task continual learning methods as illustrated in Fig. \ref{fig:ftcl}, analyzing extensively their main motivations, proposed solutions, and related evaluations.

\subsection{Regularization-based methods}
In this category, the solution is characterized by adding explicit regularization terms to balance the old and new tasks.
Bakman \textit{et al.} \cite{bakman2023federated} aimed at addressing the global catastrophic forgetting problem in FTCL under realistic assumptions that do not require access to past data samples.
The authors compared and analyzed the conventional regularization-based approaches and proposed a federated orthogonal training (FOT) framework.
FOT uses their proposed FedProject average method in the aggregation to make the global updates of new tasks orthogonal to previous tasks’ activation principal subspace to decrease the performance disruption on old tasks.
Their evaluation was compared with state-of-the-art methods, and the results indicated that FOT alleviates global forgetting while maintaining high accuracy performance with negligible extra communication and computation costs.

\subsection{Architecture-based methods}

To better solve the inter-task interference problem in FTCL, constructing specific modules or adding different parameters in the architecture is an effective and flexible solution that can explicitly help.
Wang \textit{et al.} \cite{wang2023peer} focused on the scenario where both data privacy and high-performance image reconstruction are required in multi-institutional collaborations.
The authors proposed a peer-to-peer federated continual learning network called icP2P-FL to alleviate catastrophic forgetting with reduced communication costs.
icP2P-FL uses the cyclic task-incremental continual learning mechanism across multiple institutions as the FTCL setting.
The authors also designed an intermediate controller that includes two modules, the performance assessment module (PAM) and the online determination module (ODM), to evaluate the model performance, determine the inter-institutional training order and adjust transmission costs in real time.

Chaudhary \textit{et al.} \cite{chaudhary2022federated} applied FCL in text classification to minimize catastrophic forgetting, maximize the inter-client transfer learning and minimize inter-client interference by proposing a framework called federated selective inter-client transfer (FedSeIT).
FedSeIT uses parameter decomposition methods to decompose each client’s model parameters into three different parameter sets to access task-adaptive parameters better and selectively leverage task-specific knowledge.
Specifically, the dense local base parameters capture the task-generic knowledge across clients.
Sparse task-adaptive parameters capture task-specific knowledge for each task.
Sparse mask parameters selectively utilize the global knowledge.
The authors also proposed a task selection strategy named selective inter-client transfer (SIT).
SIT is designed for efficient assessment of domain overlap at the global server using encoded data representations and selection of relevant task-adaptive parameters of foreign clients without sharing data, therefore preserving privacy while keeping the performance.
In evaluation, they used five datasets with unique labels as the FTCL scenario to demonstrate the effectiveness compared with the baseline method.

Zhang \textit{et al.} \cite{zhang2022cross} proposed a parameter decomposition-based FCL framework named Cross-FCL.
Cross-FCL uses additive parameter decomposition to separate knowledge of the local model into base parameters for common knowledge and task-specific parameters for personalized knowledge of the current local task to minimize the interference between federated learning and continual learning.
The authors also introduced cross-edge strategies on biased global aggregation and local optimization, which helps reduce memory and computation costs as well as balancing memory usage and adaptation trade-offs. 
The authors built a testbed for multi-edge federated learning on real-world image
recognition datasets and other public datasets that are divided into different disjoint sub-datasets as local task datasets in FTCL settings to demonstrate the effectiveness of the proposed Cross-FCL framework compared with the baseline.

\subsection{Replay-based methods}

In FTCL, replay-based methods include saving samples in memory and approximating and recovering old data distributions which are then used to rehearse knowledge in training current tasks.
Zizzo \textit{et al.} \cite{zizzo2022federated} defined the classic FTCL problem and proposed to mitigate the catastrophic forgetting by extending the conventional local replay methods with the global buffer to adapt a novel scenario where clients dynamically join the FL system and have varying participation rates in training rounds.
By sharing differential private data of participating clients to the global aggregator, the local models can access prior task information across clients after aggregation and use the global buffer together with the local buffer to improve performance.

Wang \textit{et al.} \cite{wang2023federated} found most existing FCL work neglected the maintenance or consolidation of old knowledge, resulting in performance degradation on previous tasks.
Therefore, the authors defined this problem under the FTCL setting and designed a federated probability memory replay (FedPMR) framework including a probability distribution alignment (PDA) module and a parameter consistency constraint (PCC) module to enhance the resistance ability to the catastrophic forgetting problem.
Specifically, PDA uses a simple but effective replay buffer in the learning process for retaining exemplars and replaying initial probability experiences from past tasks to solve the probability bias problem occurring in previous tasks.
PCC personalizes the guidance for past tasks learned at different times with the adaptive weight assignment to mitigate the imbalance in parameter variations between previous and new tasks.

Recently, to solve the challenges of continually learning new tasks without forgetting previous ones due to limited resources on edge devices for transformer-based computer vision models, Zuo \textit{et al.} \cite{zuo2024fedvit} proposed a framework for FCL of vision transformers (ViTs) on edge devices, called FedViT. 
FedViT addresses the challenges of catastrophic forgetting, negative knowledge transfer, and scalability issues in FCL under FTCL settings. 
By considering the limited storage and computation capabilities of edge devices, FedViT utilises a small number of samples from each task to improve the performance against the above challenges.
It proposes a knowledge extractor that retains critical knowledge from past tasks using a small subset of samples, a gradient restorer that converts this knowledge into gradients to help the model recover past task knowledge quickly, and a gradient integrator that ensures the combination of new and old task gradients does not lead to a loss in accuracy for any task.

\subsection{Meta Learning-based Approaches}
FTCL can be achieved not only by adding additional terms to the loss function but also by explicitly designing and integrating other optimization methods such as meta-learning.
Schur \textit{et al.} \cite{schur2023lifelong} focused on a lifelong learning (i.e., continual learning) scenario where an agent faces kernelized bandit problems sequentially, with different unknown but shared kernel information between problems.
The authors designed a lifelong bandit optimizer (LIBO) based on meta-learning approaches to transfer knowledge across Bayesian optimization problems and extended LIBO to F-LIBO under the federated learning framework.
In F-LIBO, each BO task is performed by a peer in a network without data exchanges, sequentially updating a kernel estimate to approximate the true kernel across tasks, guaranteeing optimal performance in comparison with an oracle with complete environmental knowledge over time.

Li \textit{et al.} \cite{li2023personalized} aimed to solve privacy threats in real-world task-incremental scenarios of distributed systems for biometrics.
The authors proposed a personalized FCL framework to avoid memory explosion and catastrophic forgetting in FTCL for biometrics.
This framework includes a continual task-distillation-based adaptive model-agnostic meta-learning module to retain the old knowledge and learn the knowledge transferring between incremental users.
Besides, a personalized FCL strategy, sharing global meta parameters and reserving local learnable learning rates, is designed in the framework to enhance local performance and reduce communication costs.

\subsection{Unsupervised Learning}

Paul \textit{et al.} \cite{paul2023masked} extend FedWeIT \cite{zhang2022cross} to an unsupervised continual federated learning framework called unsupervised continual federated masked autoencoders for density estimation (CONFEDMADE).
CONFEDMADE integrates masked autoencoders with federated learning within an unsupervised learning framework under the FTCL setting.
It indirectly benefits from the experiences of other clients without direct exposure to specific tasks and data to protect privacy.
The masked autoencoders with setup masking strategy integrate with task attention mechanisms, facilitating selective knowledge transfer between clients to solve catastrophic forgetting.

\subsection{Summary and analysis of FTCL approaches}
In summary, as the task identity is explicitly provided in FTCL, it will be more effective and efficient to train models with task-specific components.
Therefore, compared to the regulation-based approaches, 
architecture-based and replay-based approaches have attracted more attention in recent research, which not only prevent catastrophic forgetting but also improve the efficiency of sharing learned representations across tasks and clients.
To further improve the effectiveness of knowledge transferring and privacy preservation between FL clients, meta-learning-based approaches and unsupervised learning approaches are integrated to better solve FTCL challenges.
However, how to optimize the trade-off between the performance and computational complexity of these advanced approaches requires further attention.

\section{Federated Continual Learning Applications}
\label{sec:apps}

The previous three sections have thoroughly described the definitions, challenges, and strategies of different scenarios of FCL.
In this section, we investigate various applications empowered by FCL.
Table \ref{table:apps} shows a summary of existing application scenarios of FCL research.

\begin{table}[htbp]
\caption{Summary of Federated Continual Learning Applications}
\label{table:apps}
\begin{tabular}{p{2.5cm} c p{9.2cm}}
\hline
\textbf{\ \ Application} & \textbf{Paper} & \textbf{\qquad\qquad\qquad\qquad\quad Brief Summary} \\ \hline
\multirow{6}{2.5cm}{Intelligent Transportation System} & \cite{reddy2023deep} & ITS‑SS: Real-time prediction in the ITS.  \\ \cline{2-3} 
\multicolumn{1}{c}{} & \cite{barbieri2022decentralized}  & C-FL: Adapt to changing environments on the road. \\ \cline{2-3} 
\multicolumn{1}{c}{} & \cite{yuan2023peer}  & FedPC: Naturalistic driving action recognition for safe intelligent transportation system. \\ \cline{2-3} 
\multicolumn{1}{c}{} & \cite{guo2023icmfed}  & ICMFed: Driver distraction detection for efficient and safe intelligent transportation system. \\ \hline

\multicolumn{1}{c}{\multirow{3}{2.5cm}{Intelligent Medical Systems}} & \cite{sun2023federated} & MetaCL: A smart physiological signal classification.\\ \cline{2-3} 
\multicolumn{1}{c}{} & \cite{huang2022continual} & ICL: Automatic brain metastasis identification.\\ \cline{2-3} 
\multicolumn{1}{c}{} & \cite{guo2022federated} & A real-time medical data processing for computer-aided diagnosis.\\ \hline

\multirow{7}{2.5cm}{Internet of Things} & \cite{jin2023federated} & MENIFLD\_QoS: Intrusion detection system in resource constrained IoT. \\ \cline{2-3} 
\multicolumn{1}{c}{} & \cite{osifeko2021surveilnet} & SurveilNet: Lightweight federated IoT surveillance system with continual learning ability. \\ \cline{2-3} 
\multicolumn{1}{c}{} & \cite{qi2022collaborative} & FIL: Modulation classification in cognitive IoTs. \\ \cline{2-3}
\multicolumn{1}{c}{} & \cite{yang2022asynchronous} & FedIL: Federated continual learning framework with asynchronous training in edge networks. \\ \hline

\multirow{4}{2.5cm}{UAVs} & \cite{he2023federated} & FCL-SBLS: Intrusion detection system in UAV Networks with low computational cost. \\ \cline{2-3} 
\multicolumn{1}{c}{} & \cite{alkouz2023failure} & Failure prediction for drones at the source or intermediate nodes. \\ \hline

\multirow{4}{2.5cm}{Smart Energy} & \cite{zhang2023incremental} &  A photovoltaic power prediction to improve the power supply reliability. \\ \cline{2-3}
\multicolumn{1}{c}{} &  \cite{zhang2023online} & FLD: A fault line detection system for medium- and low-voltage power distribution networks. \\ \hline

\multirow{2}{2.5cm}{Digital Twin} &  \cite{lv2023blockchain} & BL-FCL: An efficient distributed model training framework for digital twin network. \\ \hline

\multirow{2}{2.5cm}{Financial Audit} & \cite{schreyer2022federated} & A federated continual learning framework improving the assurance of financial statements. \\ \hline

\multirow{3}{2.5cm}{Robotics} &  \cite{yu2022towards} &  A visual obstacle avoidance system for robots. \\ \cline{2-3}
\multirow{1}{2.5cm}{} &  \cite{guerdan2023federated} &  A federated continual learning framework for socially aware robotics supporting settings personalization. \\ \hline
\end{tabular}
\end{table}

\subsection{Intelligent Transportation System}

Reddy \textit{et al.} \cite{reddy2023deep} applied incremental federated learning to process the real-time data collected in the moving vehicles to achieve real-time prediction tasks in the intelligent transportation system.
The prediction models deployed in the edge node will be incrementally trained based on the vehicle learning updates after a predetermined amount of time, achieving real-time forecasting of the state of the road and other conditions for the autonomous vehicle system.

From the vehicle aspect, Barbieri \textit{et al.} \cite{barbieri2022decentralized} periodically collected new sensor data for model training to adapt to changing environments on the road. 
They considered this incremental data in the vehicle-to-everything networks and then applied continual learning settings in their decentralised consensus-driven federated learning method.

To improve the safety of ITS, Yuan \textit{et al.} \cite{yuan2023peer} utilised federated continual learning for naturalistic driving action recognition to prevent driver distraction, reduce the risk of traffic accidents, and alleviate the privacy concerns caused by in-cabin cameras.

Under a similar scenario, Guo \textit{et al.} \cite{guo2023icmfed} targeted the dynamics and heterogeneity challenges within real-world driver distraction detection and proposed a cost-efficient mechanism ICMFed by integrating incremental learning, meta-learning and federated learning to improve the efficiency and safety of intelligent transportation systems.

\subsection{Intelligent Medical Systems}
Sun \textit{et al.} \cite{sun2023federated} combined federated learning, meta-learning-empowered continual learning and block-chain for physiological signal classification to protect data privacy and overcome the catastrophic forgetting problem.
In their proposed framework, the federated learning method was used to train an Auto-Encoder-based feature extractor for the original physiological signal. 
They proposed a knowledge base module to process and store the knowledge representations learned by each task to solve the catastrophic forgetting caused by time, domain and institution change.
Specifically, they created a mask function for each task using the feature representation vectors obtained by the feature extractor and used meta-learning methods to continuously accumulate important knowledge of all tasks in updating the knowledge base. 

Brain metastasis identification is a critical scenario that requires multicenter collaboration while maintaining strict data privacy requirements among involved medical institutions.
Huang \textit{et al.} \cite{huang2022continual} proposed an effective continual learning method integrated with peer-to-peer federated learning to address the performance fluctuation in cyclic weight transfer.
They investigated regularization-based methods and utilised synaptic intelligence by adding penalties for important network parameter changes.
This method can effectively improve automatic brain metastasis identification sensitivity with peer-to-peer federated learning.

Computer-aided diagnosis (CAD) is a critical research in the medical field, which has benefited from advances in AI technology in recent years.
To further achieve real-time medical data processing and human-like progressive learning, Guo \textit{et al.} \cite{guo2022federated} combined the idea of federated learning and incremental learning and proposed a real-time medical data processing method, which reduces the time and space resource costs while mitigates the catastrophic forgetting of the disease diagnosis model.

\subsection{Internet of Things}

With the increasing zero-day attacks which may escape the existing intrusion detection system through unknown vulnerabilities caused by collecting sensitive information such as voice, fingerprint, and image in IoT devices, Jin \textit{et al.} \cite{jin2023federated} proposed a federated continual learning method which embeds discriminative auto-encode model to help the intrusion detection model update with changes of attacks.
Considering the resource-constrained IoT device features, their proposed method can infer unseen network attacks while performing fine-grained known attack identification without intensive model retraining.

In IoT surveillance networks, introducing federated learning to develop a collaborative surveillance system is necessary to solve the problem of data sharing and the inadequacy of training data in anomaly detection.
Osifeko \textit{et al.} \cite{osifeko2021surveilnet} presented a lightweight scheme that allows nodes to learn from new anomalies continuously.
Their proposed SurveilNet updates the model after receiving a false classification report or interval trigger and then slows down the learning process from the obtained knowledge to prevent forgetting issues.

In the cognitive Internet of Things, modulation classification is an essential enabler for primary user detection and signal recognition.
To process a large amount of heterogeneously cognitive IoT data in a distributed mechanism, Qi \textit{et al.} \cite{qi2022collaborative} proposed a federated continual learning method with knowledge distillation to learn the modulation classification knowledge of private classes in each local device. 
Similar to \cite{mori2022continual}, they divided the training into two phases, i.e., warm-up phase for global model learning and customised incremental learning phase for client model learning.

Yang \textit{et al.} \cite{yang2022asynchronous} proposed a federated continual learning framework with an asynchronous semi-supervised training algorithm.
Their proposed FedIL framework can help open platform applications such as IoT to prevent deep learning models from forgetting the learned information of labelled data and accelerate the convergence of the global model during training.

\subsection{UAVs}

He \textit{et al.} \cite{he2023federated} combined a stacked board learning system with federated continual learning to accommodate the increment of input data and enhancement nodes in UAV systems.
Their proposed model can effectively relieve the catastrophic forgetting problem generated by dynamic data collection, and improve the accuracy of intrusion detection with low computational cost.

The failure detection is also an essential module in UAV networks for swarm-based drone delivery services.
To efficiently utilise the energy of UAVs and the knowledge learned from old drone flight history, Alkouz \textit{et al.} \cite{alkouz2023failure} proposed a weighted continual federated learning method by allocating different weights to balance the importance between old and new flying data of drones incrementally, which performs the failure prediction at the source or when the drones land at intermediate nodes.

\subsection{Smart Energy}
Solar energy has become the most potential alternative energy source because of its inexhaustibility and non-pollution.
However, photovoltaic (PV) power output has strong fluctuation and intermittency, and its power curve has obvious non-stationarity. As a result, PV power generation is bound to affect the power quality and supply-demand balance when connected to the grid on a large scale.
In this scenario, Zhang \textit{et al.} \cite{zhang2023incremental} proposed a federated continual learning method which utilised the broad learning system through regional data sharing with incremental model update strategy to predict PV data to ensure the power supply reliability.

The power transmission and distribution networks with voltage levels of 35 kV and below are treated as critical infrastructures for large-scale power supply in cities and industrial areas.
However, achieving accurate fault line detection is challenging in medium- and low-voltage distribution network systems because of training data scarcity caused by inactive relay protection devices under this scenario.
To solve this problem, Zhang \textit{et al.} \cite{zhang2023online} proposed a fault line detection system by integrating federated learning and incremental learning strategy to improve the detection accuracy for small-sample and streaming data environments.

\subsection{Digital Twin}
In the digital twin network (DTN), distributed data protection mechanisms are considered to be utilised to mitigate user privacy threats.
To achieve this objective, Lv \textit{et al.} \cite{lv2023blockchain} improved the federated learning framework with continual learning and proposed a blockchain-based secure distributed data sharing architecture.
In this work, their proposed architecture can avoid retraining when new data comes by introducing Broad Learning into Federated Continuous Learning to speed up the model training process in DTN.

\subsection{Financial Auditing}
Some researchers are also exploring the improvement of financial statements through auditing.
Schreyer \textit{et al.} \cite{schreyer2022federated} identified two data distribution shift problems as catastrophic forgetting and model interference during auditing.
To solve these two problems, they proposed a federated continual learning framework which applied an auto-encoder network model to utilise the previous knowledge.
Their proposed framework enables auditors to incrementally learn industry-specific models from distributed data of multiple audit clients to improve the assurance of financial statements.

\subsection{Robotics}
Obstacle avoidance is a critical and essential function in autonomous mobile robot development. 
Robots need to have the capabilities of continually learning the model for obstacle avoidance like humans.
Yu \textit{et al.} \cite{yu2022towards} proposed a federated continual learning-empowered obstacle avoidance covering data collection, model training, and model sharing.

In the domain of socially aware robotics, Guerdan et al. \cite{guerdan2023federated} proposed a framework that enables robots to personalize their settings for new individuals or groups based on FCL. 
They introduced four key components as evaluation metrics for the decentralized robot learning framework: adaptation quality, adaptation time, knowledge sharing, and model overhead. 
Moreover, they developed an Elastic Transfer method based on importance regularization, which facilitates retaining relevant parameters across multiple robots, thereby enhancing knowledge sharing among robots and improving both the quality and speed of adaptation.

\section{Future Directions and Challenges}
\label{sec:future}
In this section, we highlight and discuss three future directions of FCL that provide promising opportunities for the growth of Edge-AI, which include: explainable FCL, algorithm-hardware co-design for FCL, and FCL with foundation models.

\subsection{FCL Benchmark}
As researches in FCL intensify, establishing a benchmark with representative datasets, fair evaluation criteria and framework is crucial for assessing existing FCL methods and guiding future developments. 
To the best of our knowledge, despite there have been some notable works \cite{dong2022federated, usmanova2021distillation, ma2022continual, yoon2021federated}, a robust and universally accepted benchmark is yet to emerge in FCL research. Moreover, existing FCL studies often adopt datasets and evaluation metrics from FL and CL fields. 
Here, we describe potential FCL benchmarks from three distinct aspects as follows.

\textbf{1) Common datasets}. A prevalent benchmark in FL is LEAF \cite{caldas2019leafa}, which comprises four vision task datasets and two for NLP tasks. In addition, specialized benchmarks are emerging for areas like multimodal federated learning \cite{feng2023fedmultimodal} and federated graph learning \cite{he2021fedgraphnn}. In CL, several classification datasets such as CORe50 \cite{lomonaco2017core50}, SHVN \cite{netzer2011reading}, Stream-51 \cite{roady2020stream51}, CUB-200 \cite{wahcaltechucsd}, CLEAR \cite{linclear}, and SlimageNet \cite{antoniou2020defining} have been specifically designed, alongside commonly used datasets like MNIST \cite{deng2012mnist}, CIFAR-100 \cite{krizhevsky2009learninga}, and ImageNet \cite{deng2009imagenet} variants. However, we believe that FCL tasks with \emph{blurry} task boundaries, where scenarios featuring class overlap or sharing across tasks, align more closely with real-world applications like e-commerce services and food image classification than the prevalent disjoint tasks \cite{bang2021rainbow, raghavan2024online}, as evidenced in the works \cite{aljundi2019gradient, koh2021online, prabhu2020gdumb, cossu2022classincrementala,hemati2023classincrementala}. Therefore, this practical setting merits increased attention in FCL studies.

\textbf{2) Diverse evaluation metrics}. Comprehensive yet unified metrics are equally important for fair comparisons in FCL experiments. Most FCL studies have concentrated on addressing the catastrophic forgetting problem by evaluating the performance of the final global or local model in FL, and performance across current, past, and future tasks in CL. The specific metrics vary, with some studies using averaged accuracy to assess forgetting \cite{dong2022federated,ma2022continual} and others employing forward and backward transfer metrics \cite{usmanova2021distillation,zhang2023target, yoon2021federated}. Apart from forgetting, future FCL research should consider the inherent variability across clients more thoroughly, particularly concerning constraints in computational capacity, energy and memory. Additionally, the characteristics of data resources, including non-i.i.d data distribution, sample quantity and class imbalance, demand significant attention, especially in the context of Edge AI. Consequently, the formulation of diverse metrics, meticulously designed to encapsulate these specific client-side factors, is indispensable for the nuanced evaluation and advancement of FCL.

\textbf{3) User-friendly and modular frameworks}. Existing frameworks and libraries such as FATE \cite{liufate}, PySyft \cite{ziller2021pysyft}, TFF \cite{tensorflow}, Flower \cite{beutel2022flowera}, FedML \cite{he2020fedmla}, FederatedScope \cite{xie2022federatedscope} and Avalanche \cite{lomonaco2021avalanche} have significantly facilitated FL and CL research. All of these tools are open-source,  accompanied by comprehensive documentation, and support effortless, customized modular implementation in practice, owing to their plug-and-play nature. Nevertheless, we firmly believe that crafting a user-friendly and modular framework stands as a fundamental and advantageous initiative to foster the FCL community for collaborative and sustainable growth. To this end, it is more efficient to introduce CL-empowered and FL-enabled plug-ins for existing FL and CL frameworks, respectively, rather than starting from scratch. Alternatively, developing a streamlined and lightweight framework dedicated to FCL presents another viable strategy.

\subsection{Explainable FCL}
For Edge-AI, intelligent models are distributedly deployed across various places such as industries or private communities with distinct security requirements and constraints.
In the near future, with the increasing demands for secure, robust and reliable FCL systems and applications, developing methods to provide insights into model updates and decisions in decentralised and collaborative FCL environments will be a trending need in the field of Edge AI.
By enhancing the explainability and interpretability of FCL for Edge AI, it is easier to detect and prevent attacks, as well as to verify and validate the correctness and fairness of the models \cite{10220797}, thereby building trust in the entire process of FCL.

Recently, some researchers have invested their efforts in combining FL with explainable AI models to enhance transparency and trustworthiness \cite{ungersbock2023explainable, huong2022federated, chen2022evfl, barcena2023enabling, holzinger2023human, pedrycz2021design}.
For example, 
the accuracy and trustworthiness of Quality of Experience (QoE) forecasting in Beyond 5G/6G network are improved in \cite{barcena2023enabling}, 
the trust and the prediction performance of anomaly detection for industrial control systems are improved by the proposed FL-based explainable model in \cite{huong2022federated}.
Meanwhile, the preliminary exploration of the interpretable CL approach was introduced in \cite{rymarczyk2023icicle}, which mitigates the interpretability concept drift and outperforms the performance of existing exemplar-free methods of common CCL.
However, they only analysed the exemplar-free scenario and closed-set recognition where training and test samples share the same label space, while not investigating the potential impact of incorporating a replay buffer on the model performance and exploring its compatibility with open-set settings where test samples do not come from the training.

There has been no research yet to explore explainable FCL, but the above requirements to improve model transparency and trustworthiness, the limitations and challenges of explainable FL and explainable CL also exist in FCL.
To solve these challenges, two important research aspects deserve attention for explainable FCL.

\textbf{1) Synergistic Consolidation.}
Simply introducing the CL methods into existing explainable FL or vice versa is not appropriate due to 
(i) the explainable FL model may present fewer capabilities in preventing catastrophic forgetting and maintaining the explainability over time to generate explanations for dynamically changing models after applying CL,
or
(ii) the explainable CL model may lack a clear understanding of how clients' contributions affect the global model considering complex aggregation mechanisms under a decentralised framework.
By further exploring the synergistic consolidation between explainable FL and CL, we can enable more effective, secure, transparent and trustworthy FCL model development. 
This synergistic consolidation for explainable FCL has the potential to facilitate the deployment of secure Edge-AI systems that are not only powerful but also ethically responsible.

\textbf{2) Scalability.}
The scalability challenge will also arise when more clients continually participate during federated training, making the process more heterogeneous and less efficient, causing the global model an increasing challenge to accurately generate and efficiently communicate meaningful and consistent explanations.
Therefore, solving scalability challenges in enhancing explainability for FCL in large-scale scenarios in Edge AI also needs to be further explored.

In short, explainable FCL will be an essential, challenging, but highly rewarding research direction, helping to accelerate the development of various robust and reliable applications of Edge AI.

\subsection{Algorithm-Hardware Co-design for FCL} 
Recently, there have been some studies investigating the possibility of the algorithm and hardware co-design in FL and CL, respectively. 
In FL, existing researches focus on addressing the computational bottleneck of cryptographic algorithms  \cite{yang2020fpga,zhang2023flash,wang2022pipefl,li2023flairs,che2023unifl}. 
In CL, the researchers emphasize fitting the computational and storage resource constraints of edge devices and accelerating the training and inference of neural networks with resistance to forgetting  \cite{bianchi2020bio,piyasena2020dynamically,piyasena2021accelerating,karunaratne2022memory,otero2023evolutionary,aggarwal2023chameleon,kudithipudi2023design}. 
However, current FCL research focuses on algorithms, while hardware-involved co-design has not been explored yet.

As network model parameters continuously increase and computational complexity significantly grows, FCL research confronts communication and computation challenges, particularly in resource-constrained environments. 
The algorithms and hardware in FCL are closely related and complement each other.
Currently, ongoing studies are further advancing the co-design approach, offering significant potential for addressing these challenges.
In this section, four potential research directions and associated challenges are outlined below.

1) \textbf{Hardware-aware pruning and quantization algorithms} \cite{hoefler2021sparsity,lin2022device} could significantly reduce communication and computational overheads in FCL. 
Furthermore, the quantized fixed-point numbers are well-suited for parallel computation on hardware such as GPU, FPGA, ASIC, and CIM. The challenge lies in maintaining the accuracy of the network model after lightweight.

2) \textbf{Neural Architecture Search (NAS)} could be used for automated hardware-aware design to address device heterogeneity in FCL \cite{he2021fednas,zhu2021real}. 
Hardware-aware NAS enables aggressive control of hardware resource requirements to ensure latency of training and inference on different devices. 
The complexity of NAS is a key barrier to the application of this method.

3) \textbf{Spare matrix multiplication and mixed precision supported hardware} could be designed to improve the performance and reduce the energy consumption of the FCL device, addressing the sparsity and low bit-width of the network structure brought by the pruning and quantization algorithms. 
However, designing hardware to support these operations and ensuring high utilization of hardware computing arrays is not trivial.

4) \textbf{Domain-specific hardware} \cite{zhang2023flash,luk2023heterogeneous} consisting of reconfigurable hardware and instructions can take advantage of its flexibility to adapt to multi-tasking in FCL, where different tasks can be abstracted into reusable basic operators for acceleration hardware design. 
The primary difficulty originates from co-optimization of the reconfigurable circuits and compilers.

Although there are still challenges in the promising research directions mentioned above, FCL for algorithm and hardware co-design is expected to become a trending topic and will fundamentally facilitate the development of intelligent learning systems.

\subsection{FCL with Foundation Models}
Foundation models (FMs), such as GPT \cite{brown2020language}, BERT \cite{devlin2018bert}, and CLIP \cite{radford2021learning}, capture rich knowledge and data representations through pre-training on large-scale datasets, making them adaptable to a wide range of downstream tasks via fine-tuning. 
It has evolved into fundamental infrastructures across domains like natural language processing (NLP), computer vision (CV), and speech processing. 
Recently, there has been an emerging trend in research that integrates FMs with CL or FL. 
For instance, the pre-trained model based on Transformers can effectively mitigate the catastrophic forgetting problem of CL compared to the convolutional neural network (CNN)-based approach in CV \cite{qu2022rethinking}. 
Similarly, FMs combined with FL also improve performance while preserving privacy in NLP \cite{tian2022fedbert}.
To further drive advancements in the combination between FCL and FMs for Edge-AI, two promising research directions are highlighted as follows.

\textbf{1) Reducing computational overhead on resource-constrained edge devices.}
FMs encounter significant challenges in communication and computation while facilitating FCL. 
Currently, the communication overhead has been mitigated by adopting parameters-efficient fine-tuning (PEFT) methods \cite{houlsby2019parameter,lester2021power,hu2021lora}, but the challenge of computational bottlenecks has not yet received sufficient attention, especially on resource-constrained edge devices in Edge-AI. 
Consequently, a promising research direction lies in reducing the computation and storage requirements of FCL with FMs by leveraging model compression and knowledge distillation techniques, making it suitable for resource-constrained devices while maintaining performance in FCL for Edge-AI.

\textbf{2) FMs-based cross-modal FCL for evolving environments.}
Although FMs hold promise for dealing with data heterogeneity, device heterogeneity, and multi-tasking in FCL \cite{babakniya2023slora,yi2023fedlora,su2023fedra, zhang2023fedyolo,zhao2023multi}, the aspect of continuous learning for data streams in dynamic settings remains underexplored. 
Additionally, the complexity of integrating multimodal data within FMs further introduces additional difficulties. 
Therefore, another promising research direction leads to the development of cross-modal FCL strategies based on FMs, aiming to adapt to different types of data drifts in evolving environments.

To summarize, the challenges and opportunities of applying FMs in FCL are intertwined. 
Integrating FCL with foundation models covers core issues of FL and DL, as well as bridging interdisciplinary fields such as data privacy, communication technology, and software engineering. 
Solving these complex problems requires a collaborative effort from both academia and industry.

\section{Conclusion}
\label{sec:con}
Edge-AI is an emerging and rapidly developing area. 
To ensure the performance of Edge-AI applications when handling various devices and evolving data at the edge, federated continual learning emerges to provide sustained adaptability and stable performance for learning models over time.
In this paper, we are the first to conduct an extensive and comprehensive survey on federated continual learning for Edge-AI and categorize three scenarios for federated continual learning based on different task characteristics: federated class continual learning, federated domain continual learning, and federated task continual learning.
We thoroughly summarised the background, challenges, problem formalisation, advanced solutions, and limitations of each scenario.
We also provide a review and summary of nine real-world applications empowered by federated continual learning
In addition, we highlighted four open research challenges and proposed prospective directions.
We hope this survey will inspire the research community to accelerate the progress of improving federated continual learning for Edge-AI.



\bibliographystyle{unsrt}
\bibliography{FCL_Survey}

\end{document}